\documentclass[review]{elsarticle}

\usepackage{lineno,hyperref}
\modulolinenumbers[5]

\usepackage[labelfont=bf,labelsep=newline,singlelinecheck=true,font=small]{
caption}
\usepackage{pdflscape}
\usepackage{subcaption}
\usepackage{graphicx}
\usepackage[dvipsnames]{xcolor}
\usepackage{amsmath,bm}
\usepackage{threeparttable}
\usepackage{multirow}
\usepackage[normalem]{ulem}
\useunder{\uline}{\ul}{}

\makeatletter
\def\ps@pprintTitle{%
  \let\@oddhead\@empty
  \let\@evenhead\@empty
  \def\@oddfoot{\reset@font\hfil\thepage\hfil}
  \let\@evenfoot\@oddfoot
}
\makeatother

\biboptions{authoryear}

\begin{document}

\begin{frontmatter}

\title{Bag of Tricks for Long-Tail Visual Recognition of Animal Species in Camera-Trap Images}

\author{Fagner Cunha\corref{mycorrespondingauthor}}
\cortext[mycorrespondingauthor]{Corresponding author}
\ead{fagner.cunha@icomp.ufam.edu.br}
\author{Eulanda M. dos Santos}
\author{Juan G. Colonna}
\address{Federal University of Amazonas, Manaus, Amazonas, Brazil}

\begin{abstract}
Camera traps are a method for monitoring wildlife and they collect a large
number of pictures. The number of images collected of each species usually 
follows a long-tail distribution, i.e., a few classes have a large number of
instances, while a lot of species have just a small percentage. Although in most
cases these rare species are the ones of interest to ecologists, they are often
neglected when using deep-learning models because these models require a large
number of images for the training. In this work, a simple and effective 
framework called Square-Root Sampling Branch (SSB) is proposed, which combines 
two classification branches that are trained using square-root sampling and 
instance sampling to improve long-tail visual recognition, and this is compared 
to state-of-the-art methods for handling this task: square-root sampling, 
class-balanced focal loss, and balanced group softmax. To achieve a more general 
conclusion, the methods for handling long-tail visual recognition were 
systematically evaluated in four families of computer vision models (ResNet, 
MobileNetV3, EfficientNetV2, and Swin Transformer) and four camera-trap datasets
with different characteristics. Initially, a robust baseline with the most 
recent training tricks was prepared and, then, the methods for improving long-tail recognition were applied. Our experiments show that square-root sampling 
was the method that most improved the performance for minority classes by 
around 15\%; however, this was at the cost of reducing the majority classes’ 
accuracy by at least 3\%. Our proposed framework (SSB) demonstrated itself to 
be competitive with the other methods and achieved the best or the second-best 
results for most of the cases for the tail classes; but, unlike the square-root 
sampling, the loss in the performance of the head classes was minimal, thus 
achieving the best trade-off among all the evaluated methods. Our experiments 
also show that Swin Transformer can achieve high performance for rare classes 
without applying any additional method for handling imbalance, and attains an 
overall accuracy of 88.76\% for the WCS dataset and 94.97\% for Snapshot 
Serengeti using a location-based training/test partition. Despite the 
improvement in the tail classes’ performance, our experiments highlight the 
need for better methods for handling long-tail visual recognition in camera-trap images, since state-of-the-art approaches achieve poor performance, 
especially in classes with just a few training instances.
\end{abstract}


\end{frontmatter}

\section{Introduction}

Wildlife monitoring using camera traps is a well-established, non-invasive, and
cost-effective method that has been used for decades to collect data on animals
in their daily lives. The collected data can be used to support ecological
studies on the terrestrial vertebrate community, such as species occupancy,
diversity and abundancy, community structure, and animal behavior
\citep{ahumada2013monitoring}. These cameras are usually activated by motion
sensors or at programmed time intervals to record short sequences of images or
videos \citep{he2016visual}. They are able to produce a large number of
pictures; for instance, the Snapshot Serengeti project captured 3.2 million
images between June 2010 and May 2013 over six seasons
\citep{swanson2015snapshot}, with this number growing to 7.1
million images over eleven seasons in recent years \citep{ss_lila}. However, 
processing this amount of data to extract information is expensive and 
time-consuming if done manually by experts or citizen scientists. Using the 
Snapshot Serengeti project as an example again, its volunteers donated more than 
17,000 hours of effort to manually label the 3.2 million images in the original 
dataset \citep{norouzzadeh2018automatically}. Nevertheless, recruiting such a 
large community of volunteers may be unfeasible for small-scale projects or 
those studies with a less charismatic fauna.

In the last decade, deep learning has become the dominant approach in image 
analysis and has been investigated for automatically extracting information from 
camera-trap data in various tasks, such as filtering empty images 
\citep{willi2019identifying, Cunha2021, yang2021systematic, yang2021adaptive, 
yang2021automatic}, classifying animal species \citep{villa2017towards, 
beery2018recognition, norouzzadeh2018automatically, tabak2019machine, 
willi2019identifying, schneider2020three, kutugata2021automatic, zhu2022class}, 
counting individuals \citep{norouzzadeh2018automatically, norouzzadeh2021deep}, 
and identifying behavior \citep{norouzzadeh2018automatically, 
schindler2021identification}. Meanwhile, several platforms have been developed 
to assist researchers in extracting information and speeding up this process 
using deep-learning models. Some of these platforms even allow a wider audience 
of camera-trap users to train their own models with user-friendly interfaces. 
For example, MegaDetector \citep{beery2019efficient} is a generalist model 
trained to localize animals, people, and vehicles, helping users to identify 
animals but not classifying them at the species level, and is effective in 
filtering empty images. While MegaDetector is based on the language Python, the 
Machine Learning for Wildlife Image Classification (MLWIC2) package was 
developed in R to facilitate the use of deep-learning models by biologists, 
since they are usually more familiar with this language 
\citep{tabak2020improving}. MLWIC2 provides trained models for animal species 
classification and for filtering empty images, and also allows users to train 
new models. Other platforms have also developed graphical user interfaces, such 
as Wildlife Insights \citep{ahumada2020wildlife}, Conservation AI 
\citep{chalmers2019conservation}, IFLA \citep{xi2021image}, and Animal Scanner 
\citep{yousif2019animal}, to aid researchers in extracting information.

\begin{figure*}[htp]
 \begin{center}
\includegraphics[width=\textwidth]{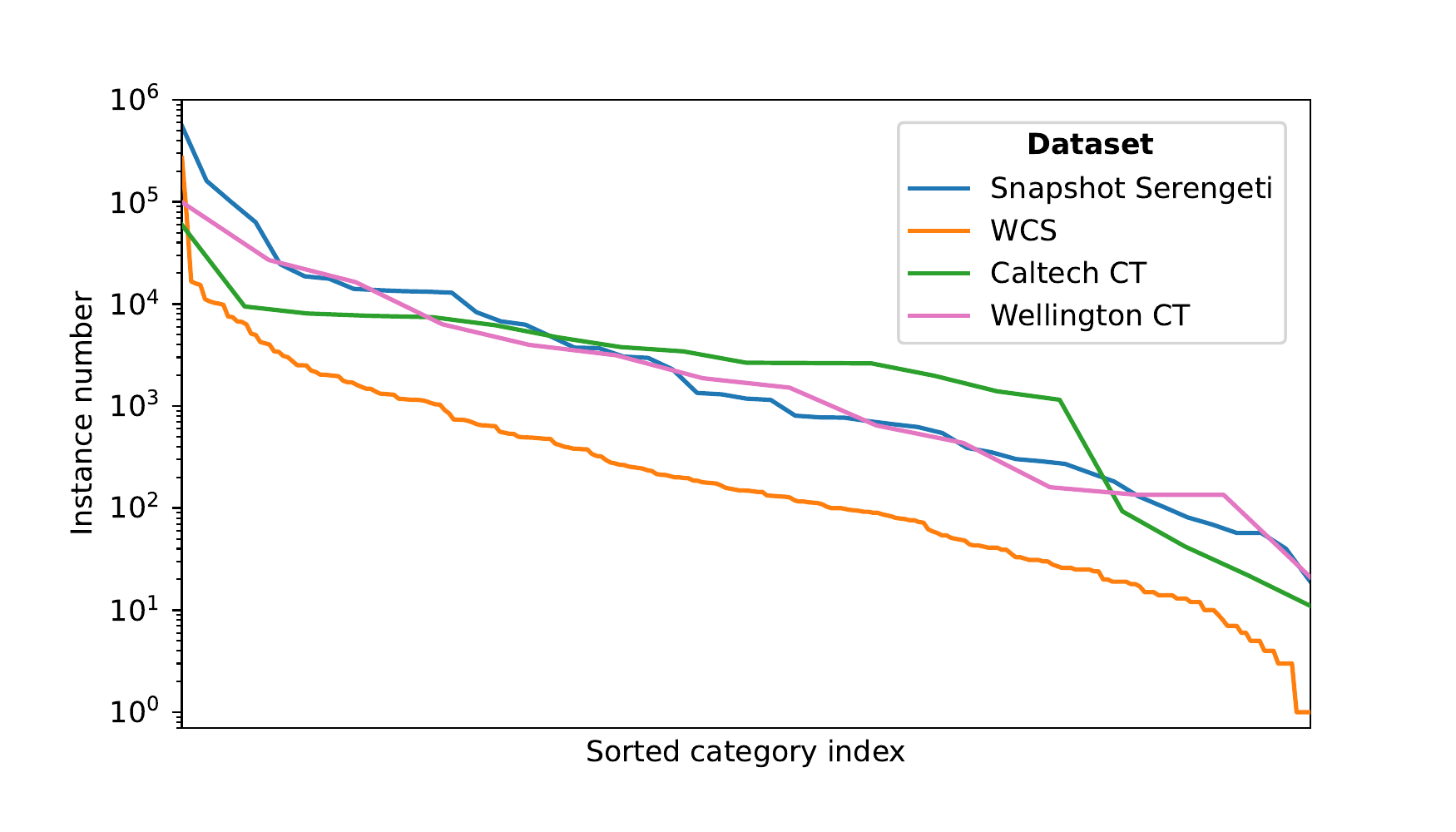}
\end{center}
\vspace{-30pt}
\caption{The number of instances per category observed in the camera-trap datasets used in this work follows a long-tail distribution. Categories are sorted by the number of instances in the training set and aligned among datasets for better visualization.}
\label{fig:train_dist}
\end{figure*}

Unlike traditional machine-learning methods, which are based on handcrafted 
features, deep-learning models automatically learn multiple levels of 
representation directly from the data \citep{bengio_deep_2012}. To achieve this, 
deep-learning models, generally deep neural networks, add more layers and more 
units within a layer, while also needing large datasets to learn these 
increasingly complex representations \citep{goodfellow2016deep}. Despite the 
large amount of camera-trap data available, extracting information from these 
images is difficult because various factors make this task hard even for humans, 
such as animals too far from the camera, camouflage, animals in a complex pose, 
and poor illumination, among others, as shown in the literature 
\citep{villa2017towards, norouzzadeh2018automatically}. Another major problem 
with camera-trap data is the extreme imbalance among categories as shown in 
\autoref{fig:train_dist}, in which some species have many images while others 
have just a few, following a long-tail distribution. As reinforced by 
\citet{schneider2020three}, models generally achieve poor recall scores for 
species with fewer training images, although these rare species are often the 
ones of greatest interest to ecologists.

Many works in the literature on machine learning have proposed techniques for 
addressing the long-tailed issue, such as re-sampling the data 
\citep{Kang2020Decoupling}, re-weighting the loss \citep{cui2019class}, or 
creating multiple branches in the networks \citep{li2020overcoming, zhou2020bbn, 
cai2021ace}. Despite this, most of the works for classifying animal species in 
camera-trap images do not apply any method for handling long-tail 
classification. Some works remove the tail classes \citep{villa2017towards, 
willi2019identifying} or group them into generic categories 
\citep{zualkernan2020towards}. For instance, \citet{villa2017towards} evaluate 
an artificially balanced version of the Snapshot Serengeti dataset, in which the 
training instance was reduced from 548,608 images to 26,000. However, this 
approach can reduce the amount of relevant information used to train the model, 
since diversity and the amount of data available for training are crucial to 
allow models to generalize well \citep{chou2020remix}. 
\citet{norouzzadeh2018automatically} apply re-weighting and over-sampling 
strategies to train a ResNet-152 model for the Snapshot Serengeti dataset. Their 
results show a 40\% improvement for some rare classes but, for other classes, 
there was a decrease in performance. \citet{schneider2020three} also used 
over-sampling combined with data augmentation to handle the extreme imbalance in 
their dataset. \citet{yang2021systematic} conducted a systematic study of class 
imbalance in the Snapshot Serengeti dataset, and evaluated various levels of 
class imbalance and their impact on model performance. However, this study 
considered only the problem of filtering empty images. Nevertheless, it is 
difficult to generalize the results presented in these works since most of the 
experiments were carried out on a single dataset. \citet{schneider2020three}, 
for example, did not reach a general conclusion, and recommended gathering more 
images for instance, which is not always feasible.

Based on the empirical observation that square-root sampling 
\citep{Kang2020Decoupling}, a type of re-sampling strategy, was the method that 
most improved the performance of the tail classes, though at the cost of 
strongly degrading the head classes, in this work a framework called Square-root 
Sampling Branch (SSB) is proposed, which combines the predictions of two 
classification branches: one trained using square-root sampling and the other 
one trained without any additional method for long-tail classification. To 
achieve more general conclusions, a systematic study of our proposed method was 
performed, as well as of other state-of-the-art methods for handling long-tail 
distribution in four camera-trap datasets -- Snapshot Serengeti, WCS, Caltech 
Camera Traps, and Wellington Camera Traps -- and four families of computer 
vision models: ResNet, EfficientNet, MobileNetV3, and Swin Transformer. A strong 
baseline training procedure, which included the most recent training 
innovations, commonly called tricks, was also prepared and was responsible for 
great improvements of classical architectures such as ResNet.

Another issue is that the high classification accuracy reported in the 
literature, such as 96.6\% acheived by \citet{norouzzadeh2018automatically} and 
98\% by \citet{tabak2019machine}, is usually not achieved when using 
deep-learning models to classify images from new locations. As shown by various 
authors \citep{beery2018recognition,tabak2019machine, schneider2020three}, it is 
difficult to reach high accuracy rates even in images from new locations in the 
same region of the images used to train the models. Due to this fact, to better 
assess the models, recent works propose to use training/test splitting based on 
locations \citep{beery2018recognition, schneider2020three}. In this case, images 
used for testing are selected from locations that are different from the 
training ones. This approach is important in the context of long-tail visual 
recognition because, if species occurrences are concentrated in a few locations, 
the model can learn spurious correlations, for example, associating the species 
to the background, which is usually static for each camera-trap node. Therefore, 
a location-based split can provide a better assessment of the generalizability 
of the models. An additional strategy is to use an object detector to localize 
animals before classifying them, thus appropriately reducing the background 
influence \citep{schneider2018deep, beery2019efficient}.

As the contributions of this paper, we can highlight:
\begin{itemize}
\item The proposal of a simple and effective framework called Square-root 
Sampling Branch (SSB) for long-tail visual recognition, which achieves a better 
trade-off as it improves performance in tail classes at the cost of head 
classes’ accuracy when compared to the state-of-the-art methods.

\item New baselines in camera-trap datasets are established using training/test 
partitioning that is based on locations by combining the most recent training 
tricks with both traditional architectures, such as ResNet-50, and newer ones 
such as Swin Transformer.

\item Our experiments show that Swin Transformer can achieve excellent results 
in camera-trap datasets even without using any additional method for handling 
the long-tail issue. 
\end{itemize}

To aid readers who may not be familiar with certain terms related to computer
science, a glossary of terms have been included in the Supplementary Materials
(see Appendix A).

\section{Materials and Methods}

In long-tail recognition problems, the classes in the datasets used to train the 
models follow a long-tail distribution, as occurs with camera-trap datasets. Our 
methodology was built to assess the effectiveness of our proposed SSB framework 
in improving the recognition performance in the tail classes and comparing it 
with methods available in the literature. First, a strong baseline was trained 
with several recent training tricks available in the literature. Then, each 
long-tail method was applied and evaluated using appropriate metrics to assess 
long-tail performance. To reach a more general conclusion, the performance of 
the methods on four long-tail camera-trap datasets with different 
characteristics and four families of computer vision models were evaluated.

In the next section, the datasets, the chosen computer vision architectures, and 
the baseline training procedure are described. Then, the families of methods for 
long-tail recognition and the selected methods used in this work are presented. 
Following this, a description of our proposed Square Root Sampling Branch (SSB) 
framework is given and, finally, the metrics used to evaluate the models are 
described.

\subsection{Datasets}

This section describes the four camera-trap datasets used in this study: 
Snapshot Serengeti, Caltech Camera Traps, WCS Camera Traps, and Wellington 
Camera Traps. These datasets were obtained from different regions around the 
globe and have different characteristics, such as the number of training 
instances, imbalance factor, and available species. While annotations for these
datasets are provided for bursts (i.e., a sequence of images), we assign
each image the label inherited from its burst annotations during the training
process. However, this approach may result in mislabeling, especially when the
animal appears in only a few images within a burst. Since manual verification
was not feasible due to the large amount of images in the datasets, the data
was used as provided, as has been done in prior studies \citep{villa2017towards,
norouzzadeh2018automatically}. Nonetheless, it is important to note that the
recommended approach is to have annotations at an image-level, as this provides
more accurate training data. Following the recent recommendations, the datasets
were split into training/validation/test partitions based on camera locations to
better assess the models’ generalization. Additionally, images were removed from
all classes which, due to the location-based split, do not have instances in the
training or test partitions. Images containing more than one species and also
corrupted pictures were also removed. \autoref{table:datasets_instances}
summarizes the number of images used from each dataset in this work. All of
these datasets follow a long-tail distribution, as shown in
\autoref{fig:train_dist}, in which a few classes (head) concentrate most of the
instances and the remaining classes have just a few samples in comparison.

\begin{table*}[htb]
\resizebox{\textwidth}{!}{
\begin{threeparttable}
\begin{center}
\caption{Number of images and locations per partition used from each dataset.}
\label{table:datasets_instances}
\begin{tabular}{llllllll}
\hline
 &  & \multicolumn{2}{c}{Training} &
\multicolumn{2}{c}{Validation} & \multicolumn{2}{c}{Test} \\ \cline{3-8}
Dataset & Classes & \# loc  & \# instances & \# loc & \# instances
 
  & \# loc   & \# instances  \\ \hline
Snapshot Serengeti & 47 & 156 & 1,059,793  & 23\textsuperscript{*}

  & 360,317   & 46\textsuperscript{*}  & 738,404
 \\
Caltech CT & 19 & 80  & 122,693   & 20\textsuperscript{*}

  & 57,162          
  & 40\textsuperscript{*}  & 61,665         \\
WCS & 247 & 2,506  & 599,040   & 313  & 79,906

  & 314 & 62,734         \\
Wellington CT & 14 & 111 & 160,937   & 36   & 49,127

  & 35  & 55,424         \\ \hline
\end{tabular}
\begin{tablenotes}
\footnotesize
\item [*] As long as there are no official test sets for Snapshot Serengeti and Caltech CT datasets, their recommended validation sets were used as the test sets in this work. For these datasets, validation sets (which were called minival) were held out from their respective training sets following the locations selected by \citet{Cunha2021}.
\end{tablenotes}
\end{center}
\end{threeparttable}}
\end{table*}

\paragraph{\textbf{Snapshot Serengeti}~\citep{dryad_5pt92}} This dataset is 
composed of more than 7 million images obtained from 225 locations spread across 
the Serengeti National Park in Tanzania, and is divided into 11 seasons. To 
maintain consistency with the literature, only the first six seasons are used in 
this work. The recommended training/validation partitions based on locations 
were used \citep{ss_lila}, with the validation set being used to evaluate the 
models. To adjust the hyperparameters, a mini-validation (minival) partition was 
held out from the training set using the locations selected by 
\citet{Cunha2021}. Specifically for this dataset, given that images from empty 
bursts (background or blank) are near identical, only one picture was picked at 
random from each empty burst.  This also helps to keep a tractable number of 
images to train the models. In total, approximately 2 million pictures 
distributed in 47 categories were used.

\paragraph{\textbf{Caltech Camera Traps}~\citep{beery2018recognition}} This 
dataset is composed of approximately of 240,000 images from 140 locations in the 
southwestern United States. The recommended location-based training/validation 
partitions were used \citep{cct_lila}, and, as for the Snapshot Serengeti, the 
validation set was used for testing, and a mini-validation partition was
created from the training set for hyperparameter adjustment based on locations
selected by \citet{Cunha2021}.

\paragraph{\textbf{WCS Camera Traps}~\citep{wcs_lila}} This dataset is provided 
by the Wildlife Conservation Society, and it is composed of 1.4 million images 
of around 675 species from 12 countries, thus making it one of the most diverse 
datasets publicly available. In our experiments, the recommended 
training/validation/test partitions based on locations were used. However, since 
some locations were not allocated to any partition, images from them were not 
used. The generic unknown categories and those not related to animals, such as 
misfire, start, and end were also manually removed. As images from classes 
without instances in training or testing sets were also removed, the resulting 
selected subset is composed of images from 247 categories.

\paragraph{\textbf{Wellington Camera Traps}~\citep{anton2018}} This dataset is 
originally composed of around 270,000 images collected from 187 camera locations 
in the Wellington region of New Zealand. The training/validation/test proposed 
by \citet{shashidhara2020sequence} was used. Images marked as unclassifiable 
were removed, in addition to images from the categories not present in the 
training or test partition, thus resulting in a dataset of around 265,000 images 
from 14 categories. Unlike the other datasets, whose empty class is the majority 
category, in this dataset the generic bird category has three times more images 
than the empty category, and has the second-largest number of instances. 

\subsection{Architectures}

Four popular computer vision architectures with different purposes and 
characteristics were chosen to assess whether long-tail classification methods 
work well on different backbones: ResNet-50 \citep{he2016deep}, 
MobileNetV3-Large \citep{howard2019searching}, 
EfficientNetV2-B2 \citep{tan2021efficientnetv2}, and Swin-S \citep{liu2021swin}. 
First, ResNet-50 was selected because it is the most common baseline for almost 
any method in computer vision. As using computer vision models directly on edge 
devices for processing camera-trap data has gained attention in the literature 
\citep{elias2017s, zualkernan2020towards, Cunha2021, zualkernan2022iot},  
MobileNetV3-Large was chosen as a benchmark for this situation. For the 
state-of-the-art families of EfficientNet \citep{tan2019efficientnet} and Swin 
Transformer \citep{liu2021swin} models, lightweight versions -- 
EfficientNetV2-B2 and Swin-S, respectively -- were chosen, since most of the 
camera-trap practitioners usually do not have access to powerful computers to 
run big models \citep{schneider2020three}. \autoref{table:arch_details} shows 
the number of model weights (without the classification layer, which varies
according to the dataset), input size, and number of FLOPs for each of these
models. FLOPs refers to the number of floating-point operations (multiply-adds)
performed by a model, and is commonly used as a measure to estimate the
theoretical computational complexity \citep{zhang2018shufflenet}. These details
can be used to compare architectures in terms of hardware requirements.

\begin{table}[htb]
\begin{center}
\caption{Details of the architectures used. The number of FLOPs (in billions)
is calculated from the literature considering the ImageNet classification head
(1,000 classes).}
\label{table:arch_details}
\begin{tabular}{lrll}
\hline
Architecture & \# weights & input size & FLOPs \\ \hline
ResNet-50   & 23,587,712   & 224 x 224  & 4.1B  \\
MobileNetV3-Large  & 4,226,432 & 224 x 224  & 0.22B  \\
EfficientNetV2-B2 & 8,769,374 & 260 x 260  & 1.7B  \\
Swin-S  & 49,173,398   & 224 x 224  & 8.7B  \\ \hline
\end{tabular}
\end{center}
\end{table}

\subsection{Baseline training settings}\label{sec:baseline_settins}

As shown recently \citep{he2019bag, bello2021revisiting, wightman2021resnet}, 
improved training strategies are responsible for significant gains in the 
accuracy of classical architectures such as ResNet-50. Following these ideas, 
several tricks were included in our baseline training procedure, and these are 
described below.

\paragraph{\textbf{Image preprocessing}} Initially, all images were resized so 
that the largest dimension was at most 768 px, and the smallest one was at least 
450 px. During training, first, a rectangular region of aspect ratio sampled in 
$[3/4, 4/3]$ and area in $[65\%, 100\%]$ was randomly cropped and resized to the 
input size of the architecture. Next, the image was flipped horizontally with a 
50\% probability. Then, RandAugment \citep{cubuk2020randaugment} was used, which 
is a method that applies a random sequence of image distortions, such as
rotation, contrast, and brightness distortions, to dynamically augment the data
during the training. This method has two hyperparameters: $N$, which is the
number of distortions applied sequentially, and $M$, which is the magnitude of
all distortions. In this work, we set $N=2$ and $M=9$ as parameters for
RandAugment. Finally, the image was rescaled to match the input scale of each
architecture, as shown in \autoref{table:arch_rescale}.
During the evaluation, the preprocessing consists only of resizing the image to 
the input architecture size and rescaling.

\begin{table*}[htb]
\caption{Input scaling procedure for each architecture.}
\label{table:arch_rescale}
\begin{center}
\begin{tabular}{lp{0.6\textwidth}}
\hline
Architecture & Scaling Procedure
                     \\ \hline
ResNet-50   & Convert from RGB to BGR, zero-center w.r.t. the ImageNet
dataset, without scaling \\
MobileNetV3-Large  & Pixels between 0 and 255 (model includes preprocessing
layer)                     \\
EfficientNetV2-B2 & Scale pixels between -1 and 1                               
                      \\
Swin-S  & Scale pixels between 0 and 1, normalize w.r.t. the ImageNet
dataset               \\ \hline
\end{tabular}
\end{center}
\end{table*}

\paragraph{\textbf{Training procedure}} The models were initialized with 
ImageNet pre-trained weights, replacing the classifier layer, and, then, were 
trained with a batch size of 64 for 30 epochs when fine-tuning the whole model 
or for 12 epochs when training only the classifier layer. The AdamW optimizer 
\citep{loshchilov2018decoupled} was employed for training with an initial 
learning rate of $10^{-5}$ and weight decay of $10^{-7}$. The learning rate was 
linearly warmed up from $0$ to the initial value for two epochs when fine-tuning 
-- one epoch when training only the classifier --, and then decayed to $0$ using 
the cosine schedule \citep{he2019bag}.

\subsection{Training tricks for long-tail visual recognition}

\begin{figure*}[htp]
 \begin{center}
\includegraphics[width=\textwidth]{
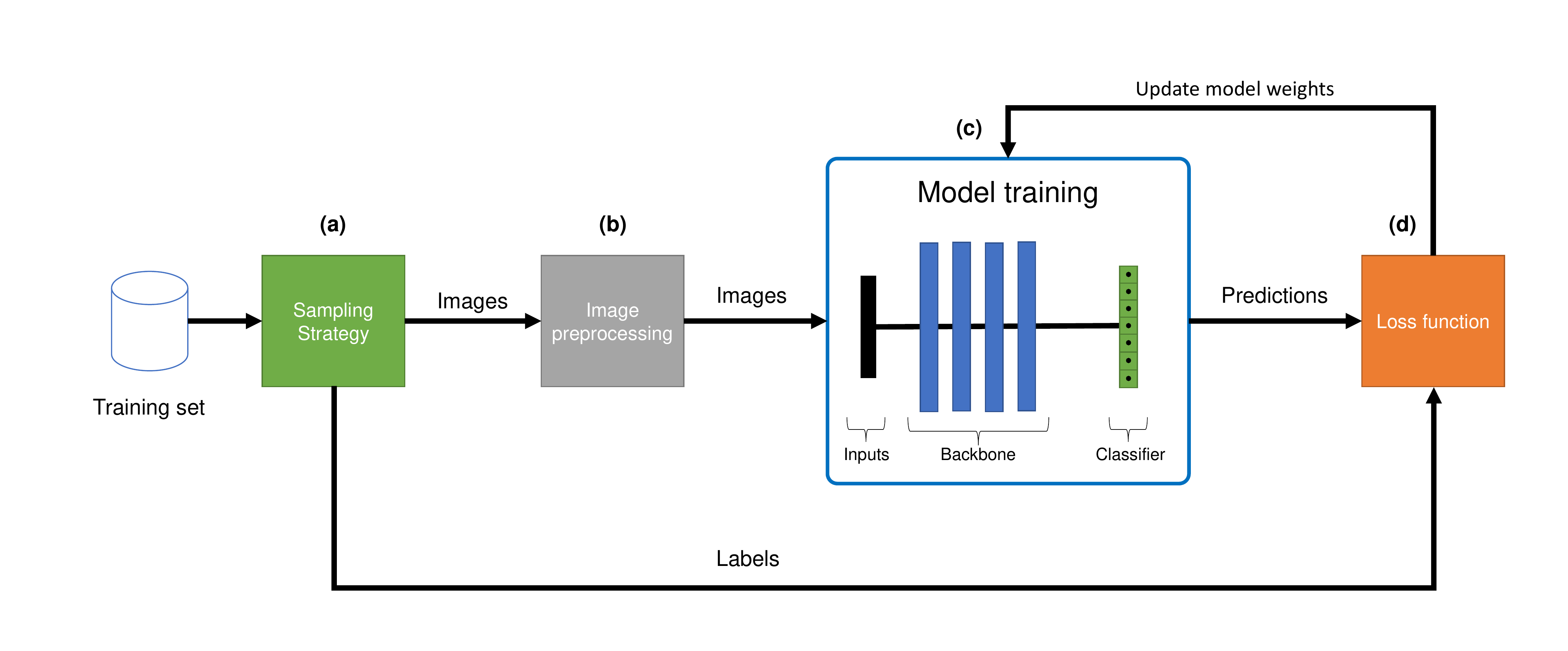}
\end{center}
\vspace{-20pt}
\caption{Schematic diagram of model training process. Each method studied in
this work operates at one or more stages: re-sampling methods change the
sampling strategy (a), re-weighting methods modify the loss function (d),
methods using multi-branch networks modify the model structure (c), and
two-stage training methods operate by selecting which model weights (c) are
trained at each stage. Our proposed SSB method operates by changing both the
sampling strategy (a) and the model structure (c). Additionally, cropping images
before classification is performed at the image preprocessing stage (b).}
\label{fig:train_diagram}
\end{figure*}

This section describes some effective methods for long-tail visual recognition 
that were evaluated in this work for camera-trap image classification. The 
methods were grouped into four main families according to their focus: 
re-sampling the data, re-weighting the loss, using two stages to focus on 
representation learning, and creating multiple branches in the network. The 
method of using an object detector to crop regions of interest from camera-trap 
images before classifying them was also included in this section. Despite not 
dealing directly with the imbalance problem, it has been shown in the literature 
\citep{schneider2018deep, beery2019efficient} that this trick improves the 
recognition score in camera-trap images. \autoref{fig:train_diagram} illustrates
a functional summary of the model training process, highlighting the stages
where each method studied in this work operates.

\subsubsection{Re-sampling}

Re-sampling methods attempt to sample the dataset in order to obtain a more 
balanced data distribution by over- or under-sampling instances according to the 
number of instances per class. On the one hand, random over-sampling consists of 
replicating instances from the minority classes; however, this method can lead 
to overfitting these instances \citep{chou2020remix, cai2021ace}. On the other 
hand, random under-sampling discards instances from the majority classes, which 
can degrade the representation learning, especially for extremely imbalanced 
datasets \citep{chou2020remix}.

In some re-sampling strategies, the probability $p_{j}$ of sampling an instance 
from class $j$ is given by:
\begin{equation}
 p_{j} = \frac{n^{q}_{j}}{\sum\limits_{i=1}^{C}n^{q}_{i}},
\end{equation}
where $C$ is the number of training classes, $n_i$ is the number of training 
instances from class $i$, and $q \in [0, 1]$. When $q=1$, each class has a 
probability $p_{j}$ directly proportional to its number of instances, i.e., each 
instance has the same probability of being sampled, for this reason, this method 
is known as instance-balanced sampling, and, in practice, it uses the original 
data distribution \citep{Kang2020Decoupling}. When $q=0$, each class has an 
equal probability ($p_j=1/C$) of being selected, this is also called 
class-balanced sampling \citep{Kang2020Decoupling, zhang2021bag}. In this work, 
following recent literature \citep{Kang2020Decoupling, zhang2021bag}, an 
alternative between those two extremes was chosen by setting $q=1/2$ -- called 
square-root sampling, since this approach returns a less imbalanced dataset.

\subsubsection{Re-weighting}

The re-weighting methods work by modifying the loss function in order to 
redefine the importance given to each class or instance during the training. 
These methods guide the network on where to pay more attention by assigning 
different weights to different categories or instances 
\citep{Kang2020Decoupling, zhang2021bag, wang2021longtailed}. One naïve approach 
is to weight the dataset based on the inverse class frequency or the inverse 
square root of class frequency\citep{chou2020remix}. More sophisticated 
approaches, such as the Focal loss and the Class-Balanced loss, have been 
developed and take into consideration factors such as how easy one sample is, or 
the effective number of samples.

Focal loss \citep{lin2017focal} was proposed in the context of the object 
detection problem, in which there is an extreme foreground-background class 
imbalance. This loss automatically down-weights well-classified samples to make 
the training focus on hard training instances. To achieve this objective, given 
a class probability prediction $p_i$ for the instance $x_i$, Focal loss 
$L_{focal}$ is defined as
\begin{equation}\label{eq:focal}
 L_{focal}=-(1-p_i)^{\gamma}\log (p_i),
\end{equation}
where $\gamma \geq 0 $ is a parameter that controls the rate at which easy 
examples are down-weighted.

The Class-Balanced loss (CB loss) \citep{cui2019class} weights each class based 
on the effective number of samples $E_n$, which is given by:
\begin{equation}
 E_{n}=\frac{1-\beta^{n}}{1-\beta},
\end{equation}
where $n$ is the number of samples and $\beta \in [0,1)$ is a hyperparameter. 
Given a class $j$ with $n_j$ training samples, the loss $L$ with respect to this 
class is weighted by a factor using the effective number of samples:
\begin{equation}\label{eq:cb}
 CB(\bm{p},j) = \frac{1}{E_{n_j}}L(\bm{p},j) = 
\frac{1-\beta}{1-\beta^{n_j}}L(\bm{p},j),
\end{equation}
where $\bm{p}$ denotes the class probabilities estimated by the model.

For representing the re-weighting methods in our experiments, the Class-Balanced 
Focal (CB-Focal) loss was used, which is the combined version of Focal loss and 
Class-Balanced loss. In this case, $L(\bm{p},j)$ in \autoref{eq:cb} is replaced 
by $L_{focal}$ from \autoref{eq:focal}. This loss weights the importance of each 
instance on the loss from the perspective of both class distribution and sample 
easiness. In our experiments, $\beta$ was set to $0.9$ and $\gamma$ to $2.0$.

\subsubsection{Two-stage training}

The extensive experiments conducted by \citet{Kang2020Decoupling} on common 
long-tail benchmarks showed that data imbalance is not an issue for learning 
high-quality representations. Their results evidenced that it is possible to 
attain a good long-tail recognition score by training only the classifier on top 
of those learned representations. Taking this into account, 
\citet{Kang2020Decoupling} propose a two-stage training method for long-tail  
recognition. In the first stage, the model is trained using the original 
unbalanced training distribution. In the second, only the classifier is trained 
using a balancing method, such as re-weighting or re-sampling.

\subsubsection{Multi-branch networks}

Methods using multi-branch networks handle the long-tail problem by creating 
expert branches to treat relatively balanced sub-groups of classes separately. 
The Bilateral-Branch Network (BNN) \citep{zhou2020bbn} uses two backbone 
branches, one is trained with a uniform sampler and the second one uses a 
sampler with a reversed distribution w.r.t. the original long-tail distribution; 
a cumulative learning strategy is then used to combine their output feature 
vectors. Ally Complementary Experts (ACE)~\citep{cai2021ace} works by using 
multiple expert branches on top of a shared backbone, in which each expert has 
individual learning blocks and a prediction layer for a diverse but overlapping 
set of categories. Then, the experts’ predictions are aggregated by averaging 
the logits that are re-scaled according to the data splits. The main idea is 
that the dominating categories (head classes) for each expert are different, and 
the overlapping categories support each other during the aggregation, especially 
for categories with few samples. Balanced Group Softmax (BAGS) 
\citep{li2020overcoming} also uses a shared backbone for representation 
learning, but the categories are divided into disjoint groups according to the 
number of instances. It was decided to test the effectiveness of BAGS because 
the winning team \citep{cunha2021iwildcam} of the iWildCam 2021 
\citep{beery2021iwildcam} -- an annual competition on extracting information 
from camera-trap images -- used it.

BAGS is trained in a two-stage fashion. First, the model is trained using a 
single softmax classification head containing all categories. Then, the
classification head is
replaced by the new groups of classification heads randomly initialized.
Finally, only these heads are trained, with all other weights frozen. Each
category $j$ is assigned to a single group $G_k$ based on its number of training 
instances $n_j$ if $s_k^l \leq n_j < s_k^h$, with $k > 0$, where $s_k^l$ and 
$s_k^h$ are the minimum and maximum number of instances for the group $G_k$ and 
$s_{k+1}^l = s_k^h$. In our experiments using BAGS, following the originally 
proposed method \citep{li2020overcoming}, the categories were split into four 
groups, with $s^l$ and $s^h$ set to $(0, 10)$, $(10, 10^2)$, $(10^2, 10^3)$, and 
$(10^3, +\infty)$, and the special group $G_0$ for foreground/background 
classification. Each group includes an extra category \textit{others} for the 
samples from classes belonging to the other groups. To balance training, the 
\textit{others} category within each batch was undersampled by a factor of 
$\beta \times n_k$, where $n_k$ is the number of instances from categories 
belonging to the group in that batch ($\beta=8$ was used). During the inference, 
a softmax is applied to each group and the probabilities are remapped back to 
the original category order and rescaled by the foreground probability. By this 
design, classes with a significantly different number of instances are isolated 
from each other during the training, thus avoiding the tail classes being 
substantially suppressed by the head classes.

\subsubsection{Cropping before classification}

As shown in \citep{schneider2018deep} and \citep{beery2019efficient}, using an 
object detector to crop regions of interest from camera-trap images before 
classifying may improve the recognition score. This training trick was included 
in our experiments to evaluate how it can enhance the species classification 
when combined with other methods used to tackle the long-tail issue. 
MegaDetectorV4 \citep{beery2019efficient} was used to generate bounding box 
predictions, which is a general-purpose animal detector based on the object 
detector Faster R-CNN~\citep{ren2015faster} and is trained on a large number of 
camera-trap images. In our experiments with cropping before classification, 
MegaDetectorV4 was run on the resized images before the random crop operation in 
the preprocessing pipeline. Only bounding boxes with a confidence score higher 
than $0.6$ were considered, otherwise, the whole image was used. Then, a squared 
crop was applied around the bounding box with the highest confidence score. 
Following this, the preprocessing procedure occurs as described in the baseline 
training settings (\autoref{sec:baseline_settins}).

\subsection{SSB: Square-root Sampling Branch}

Our proposed framework is a multi-branch network approach that was inspired by 
an empirical observation in our preliminary experiments: the square-root 
sampling strategy was the method that most improved the accuracy of the tail 
classes over the baseline; however, at the cost of significantly decreasing the 
performance of the head classes. The question to answer is: is it possible to
combine the \textbf{S}quare-root \textbf{S}ampling \textbf{B}ranch (SSB) 
predictions with the baseline to improve the tail classes with minimal 
performance degradation for the head classes?

\begin{figure*}[htp]
\vspace{-60pt}
 \begin{center}
\includegraphics[width=\textwidth]{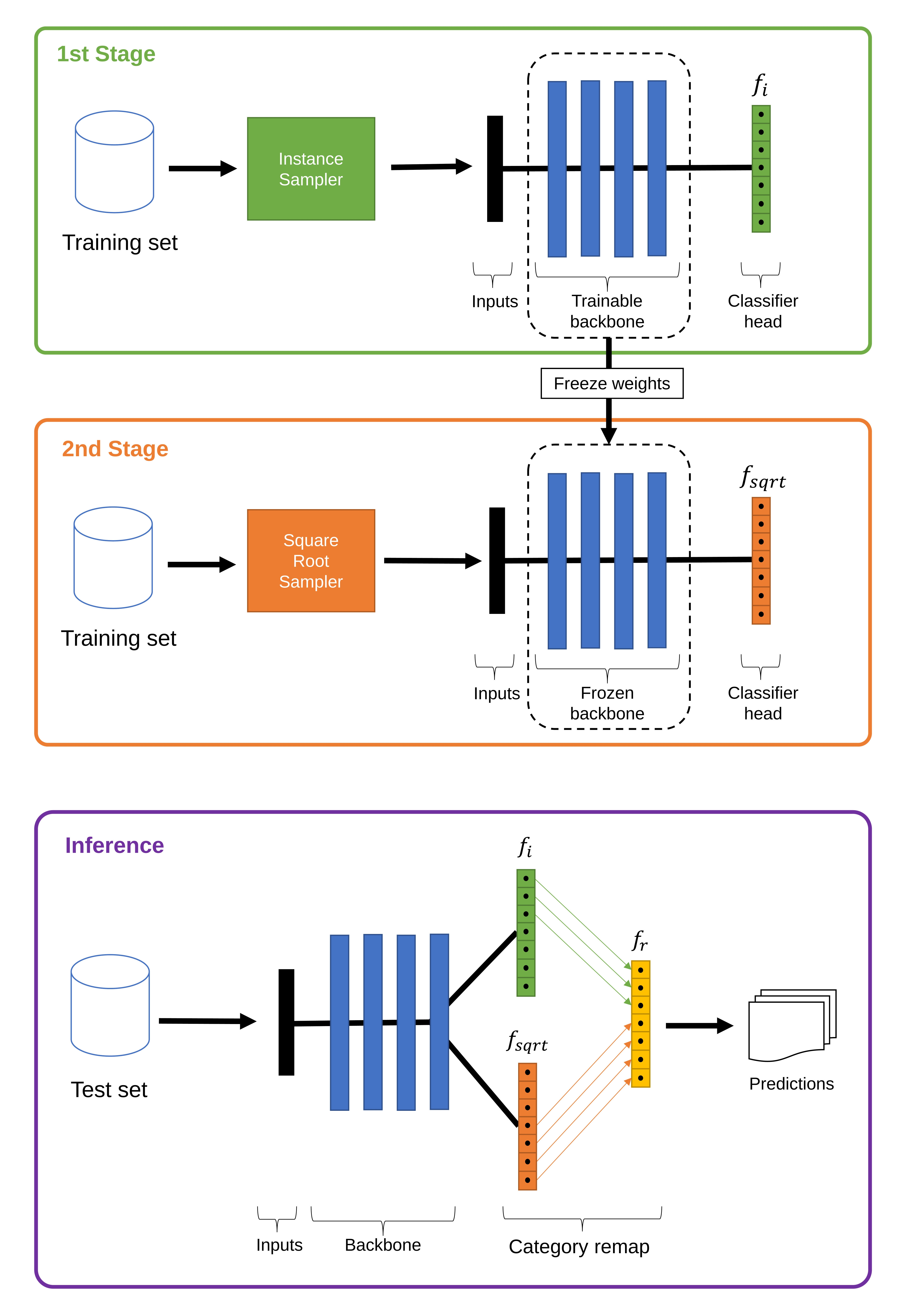}
\end{center}
\vspace{-16pt}
\caption{Framework of our Square-root Sampling Branch (SSB). It is trained in two stages. In the first stage, the classifier ($f_i$) and the backbone are jointly trained using the original data distribution. In the second stage, a new classifier ($f_{sqrt}$) is trained using square-root sampling and the backbone trained in the first stage is kept frozen. During the inference, predictions from both classifier heads are aggregated by masking the classes to keep the head classes probabilities of $f_i$ and the remaining classes probabilities of $f_{sqrt}$.}
\label{fig:ssb}
\end{figure*}

As illustrated in \autoref{fig:ssb}, the proposed framework is trained in a 
two-stage fashion. First, the classifier ($f_i$) and the backbone are trained 
jointly using the original data distribution (instance-based sampling). For the 
second stage, the network backbone is frozen and the classifier is replaced by a 
new randomly initialized one. Then, this new classifier ($f_{sqrt}$) is trained 
using data sampled by the square-root sampling strategy. Both classifiers are 
connected on top of the backbone during the inference. Their predictions are 
aggregated by masking the softmax outputs to keep only the head classes 
probabilities from the $f_i$ head and the remaining classes probabilities from 
the $f_{sqrt}$ head.

The $C$ classes were divided into four groups based on the number of training 
instances, in which a class $j$ with $n_j$ instances belongs to group $G_k$ if 
$s_k^l \leq n_j < s_k^h$. The limits of each group are set to $(0, 10)$, $(10, 
10^2)$, $(10^2, 10^3)$, and $(10^3, +\infty)$. However, SSB does not have a 
special group for the background class, with this class belonging to the group 
$G_4$. Given a feature vector $h$ generated by the backbone and the predictions 
$p_i = f_i(h)$ and $p_{sqrt} = f_{sqrt}(h)$ generated by each classification 
head with softmax, the final remapped prediction vector $p_r$ is given by:
\begin{equation}\label{eq:ssb}
 p_r = f_r(p_i, p_{sqrt}) = Qp_i + (\bm{1} - Q)p_{sqrt},
\end{equation}
where $Q$ is a matrix $C\times C$ that masks the softmax outputs based on which 
group a class belongs to:
\begin{equation}\label{eq:ssb_mask}
 Q = \left\{ q_{a,b} | q_{a, b} = \begin{cases}
      1 & \text{if } a=b \text{ and the class } a \in G_4 \text{,}\\
      0 & \text{otherwise.}
    \end{cases}\right\}.
\end{equation}
As with BAGS post-processing, the final prediction vector produced by SSB is not 
a real probability vector because its sum is not equal to 1.

\subsection{Evaluation metrics}

Traditionally, deep-learning benchmarks use top-1 and top-5 accuracies to 
evaluate model performance \citep{ILSVRC15}. However, these metrics are biased 
towards majority classes for unbalanced datasets, and do not accurately assess 
the tail classes. Therefore, following \citet{li2020overcoming}, the categories 
in the test set of each dataset were grouped into four bins according to the 
amount of training instances, and report the accuracy of each bin. $Bin_i$ 
contains the categories composed of $10^{i-1}$ to $10^i$ instances. Hence, 
$Bin_1$ and $Bin_2$ represent the tail classes, while the $Bin_3$ and $Bin_4$ 
represent the head classes.

In addition to the accuracy using bins, the macro-averaged F1-score is also 
reported, which is, according to \citet{picek2022danish}, more suitable for 
long-tail distributions observed in nature. The F1-score $F_{1_i}$ of a class 
$i$ is the harmonic mean of its precision $p_i = \frac{tp_i}{tp_i + fp_i}$ and 
recall $r_i = \frac{tp_i}{tp_i + fn_i}$. True positive ($tp_i$) is the number of 
correct predictions for class $i$, False positive ($fp_i$) is the number of 
instances incorrectly classified as belonging to $i$, and False negative 
($fn_i$) is the number of instances from class $i$ wrongly classified as not 
belonging to $i$. The macro-averaged F1-score $F_1^m$ across $C$ classes is the 
mean of the class-wise F1-score:

\begin{equation}
 F_1^m = \frac{1}{C} \sum\limits_{i=1}^{C}F_{1_i} = \frac{1}{C} 
\sum\limits_{i=1}^{C}(2 \times \frac{p_i \times 
r_i}{p_i + r_i})^.
\end{equation}

\section{Experiments and Results}
Our experiments were broken down into three main series. In the first, our 
proposed method SSB is compared to three two-stage methods for long-tail 
problems, as well as the baseline. In the second series, the same methods are 
compared, but using cropped images as inputs instead of the entire images.
Finally, in the third series, SSB is compared to single-stage training methods.
The results presented in this section were obtained using the test sets
specified for each dataset in \autoref{table:datasets_instances}.

\subsection{Comparison between SSB and two-stage methods for long-tail visual
recognition}

In the first experiment, the effectiveness of our proposed SSB method when using 
the full image as input for the training and testing stages was evaluated. As 
our proposed method is trained using a two-stage strategy, in this experiment, 
only the results with two-stage training methods were compared. Initially, each 
architecture was trained using our baseline training settings, and then each 
long-tail method was applied to the classifier layer, though keeping the 
backbone frozen during the second-stage training. This allows a fair comparison 
between the methods in terms of the representation learned by the backbone.

The accuracy results are shown in \autoref{table:fullimage_results}. The 
proposed SSB consistently improved accuracy for $Bin_1$, $Bin_2$, and $Bin_3$ 
when compared to the baseline for almost all cases, with a minimal loss in the 
head class accuracy ($Bin_4$), and achieved the best or the second-best accuracy 
per bin for most of the cases. The best results for $Bin_1$, $Bin_2$, and 
$Bin_3$ were achieved by the square-root sampling, with an improvement ranging 
from 3\% to more than 15\% over the baseline. However, such an improvement came 
at the cost of reducing the accuracy of the majority classes ($Bin_4$) by at 
least 3\%. This aspect may be observed in the overall accuracy, since the 
square-root sampling achieved the worst results among all the methods. On the 
other hand, BAGS achieved the best overall accuracy results for some cases; 
while, in others, it even reduced the accuracy for the tail classes compared to 
the baseline.  All methods, however, failed to improve the performance on tail 
classes ($Bin_2$) for the smaller datasets (Caltech and Wellington). In this
case, accuracies were close to zero.

\begin{figure*}[htp]
\centering
\vspace{-24pt}
\makebox[\textwidth][c]{
\begin{subfigure}[b]{\textwidth}
\begin{center}
\includegraphics[height=0.17\textheight]{
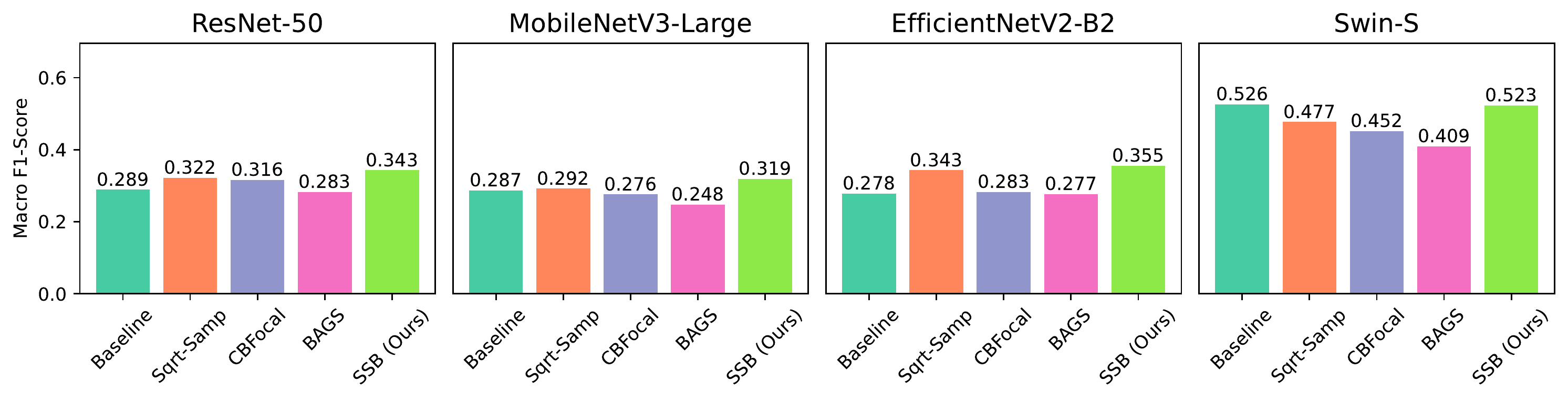}
\end{center}
\vspace{-16pt}
    \caption{WCS Camera Traps}
    \label{fig:wcs_fullimage_f1}
\end{subfigure}
}
\hfill

\begin{subfigure}[b]{\textwidth}
\begin{center}
\includegraphics[height=0.17\textheight]{
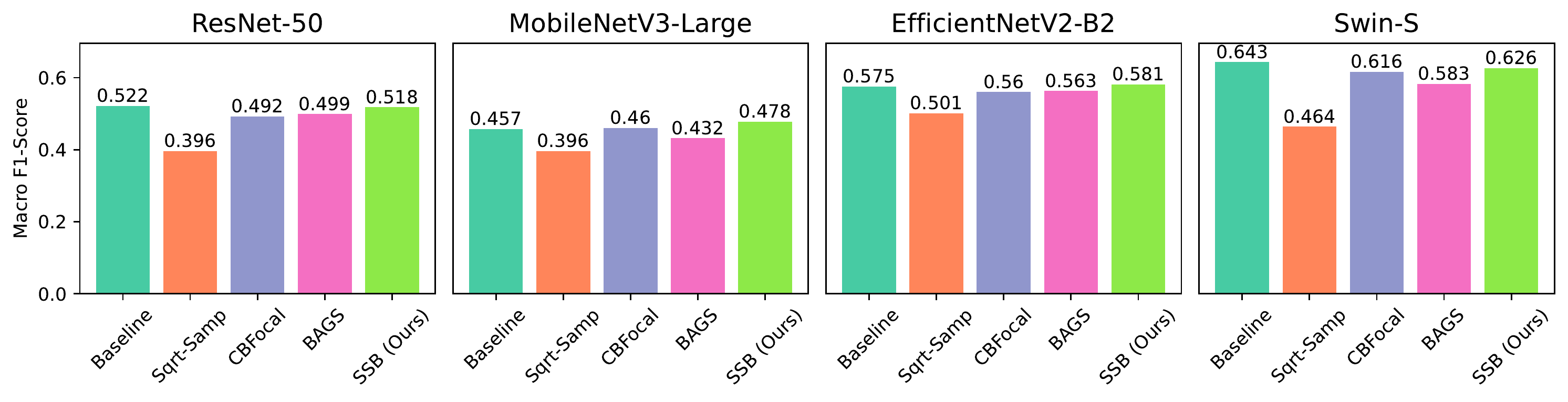}
\end{center}
\vspace{-16pt}
    \caption{Snapshot Serengeti}
    \label{fig:ss_fullimage_f1}
\end{subfigure}
\hfill

\begin{subfigure}[b]{\textwidth}
\begin{center}
\includegraphics[height=0.17\textheight]{
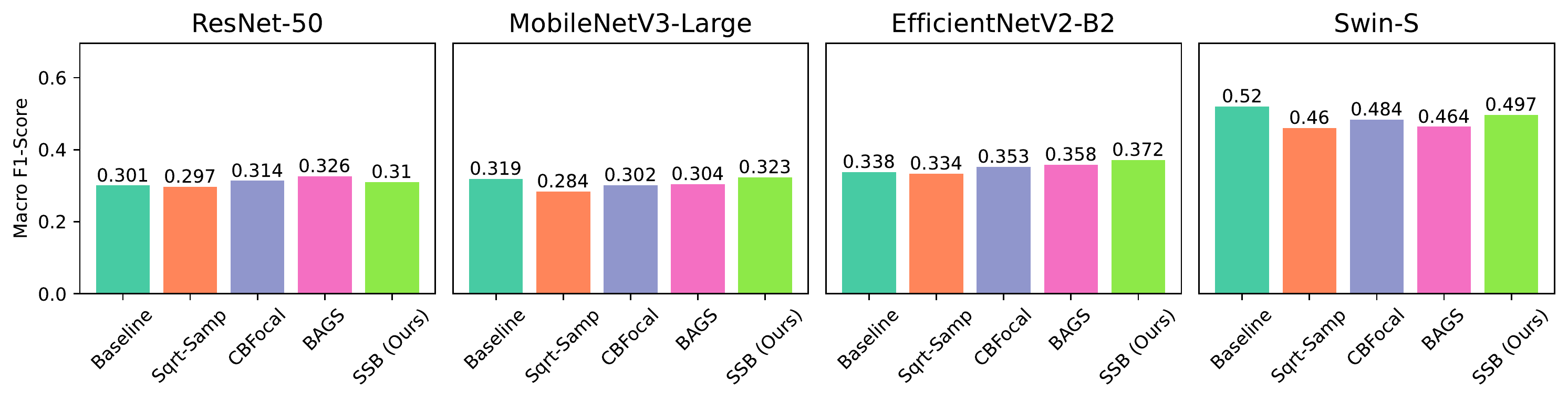}
\end{center}
\vspace{-16pt}
    \caption{Caltech Camera Traps}
    \label{fig:catech_fullimage_f1}
\end{subfigure}
\hfill

\begin{subfigure}[b]{\textwidth}
\begin{center}
\includegraphics[height=0.17\textheight]{
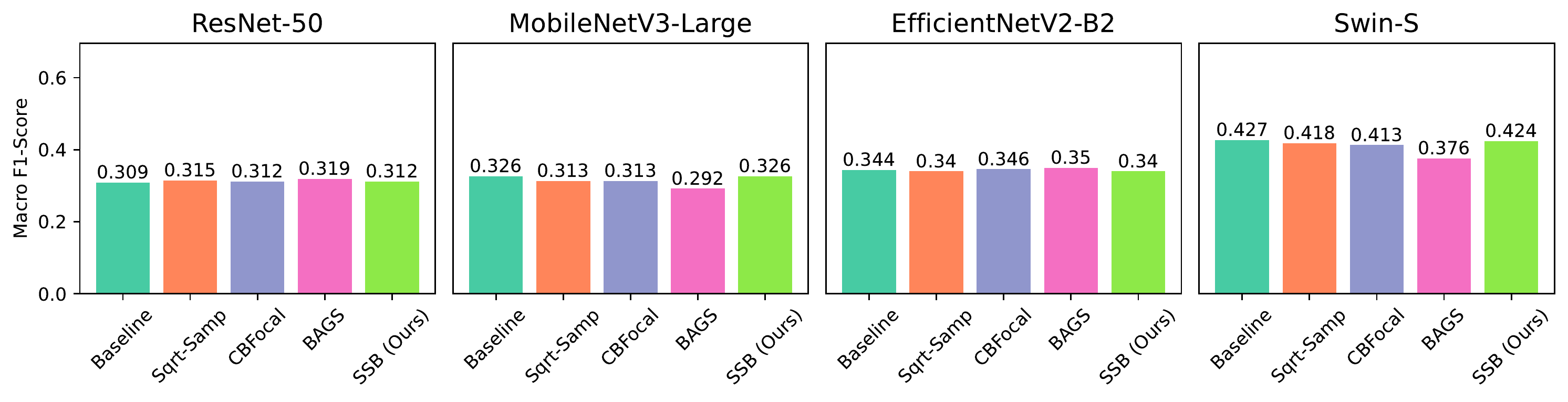}
\end{center}
\vspace{-16pt}
    \caption{Wellington Camera Traps}
    \label{fig:wellington_fullimage_f1}
\end{subfigure}

   \caption{Comparison in terms of macro-averaged F1-score for all combinations of datasets and models. SSB achieved a score that was higher than the other long-tail visual recognition methods. It is worth mentioning that for the Swin-S model, the baseline obtained the best macro F1-score for all the datasets. SSB is the second best.}
\label{fig:fullimage_f1}
\end{figure*}

\begin{figure*}[htp]
\centering
\vspace{-36pt}
\makebox[\textwidth][c]{
\begin{subfigure}[b]{\textwidth}
\begin{center}
\includegraphics[height=0.4\textheight]{
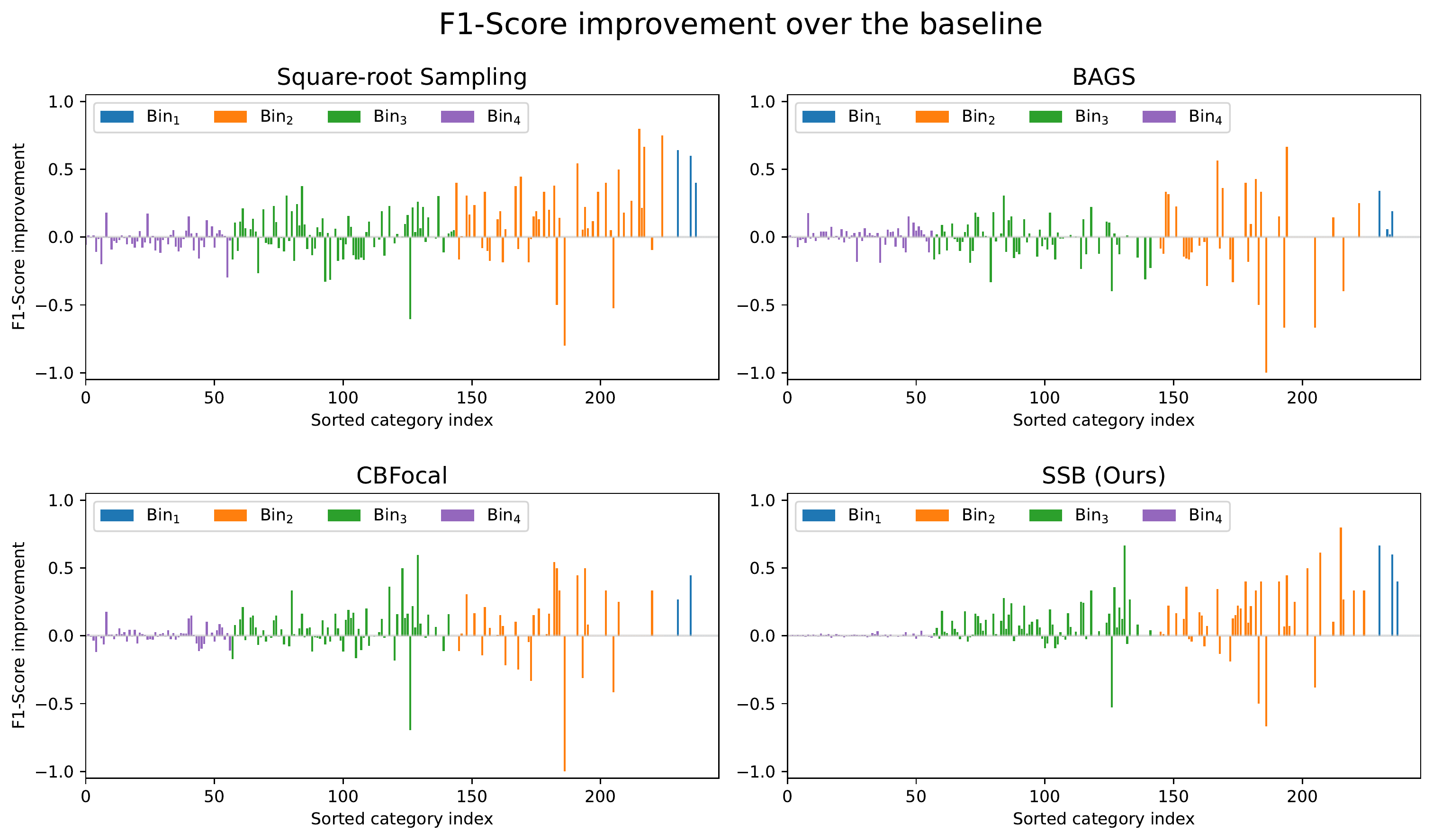}
\end{center}
\vspace{-16pt}
    \caption{ResNet-50}
    \label{fig:wcs_fullimage_f1_resnet}
\end{subfigure}
}
\hfill

\begin{subfigure}[b]{\textwidth}
\begin{center}
\includegraphics[height=0.4\textheight]{
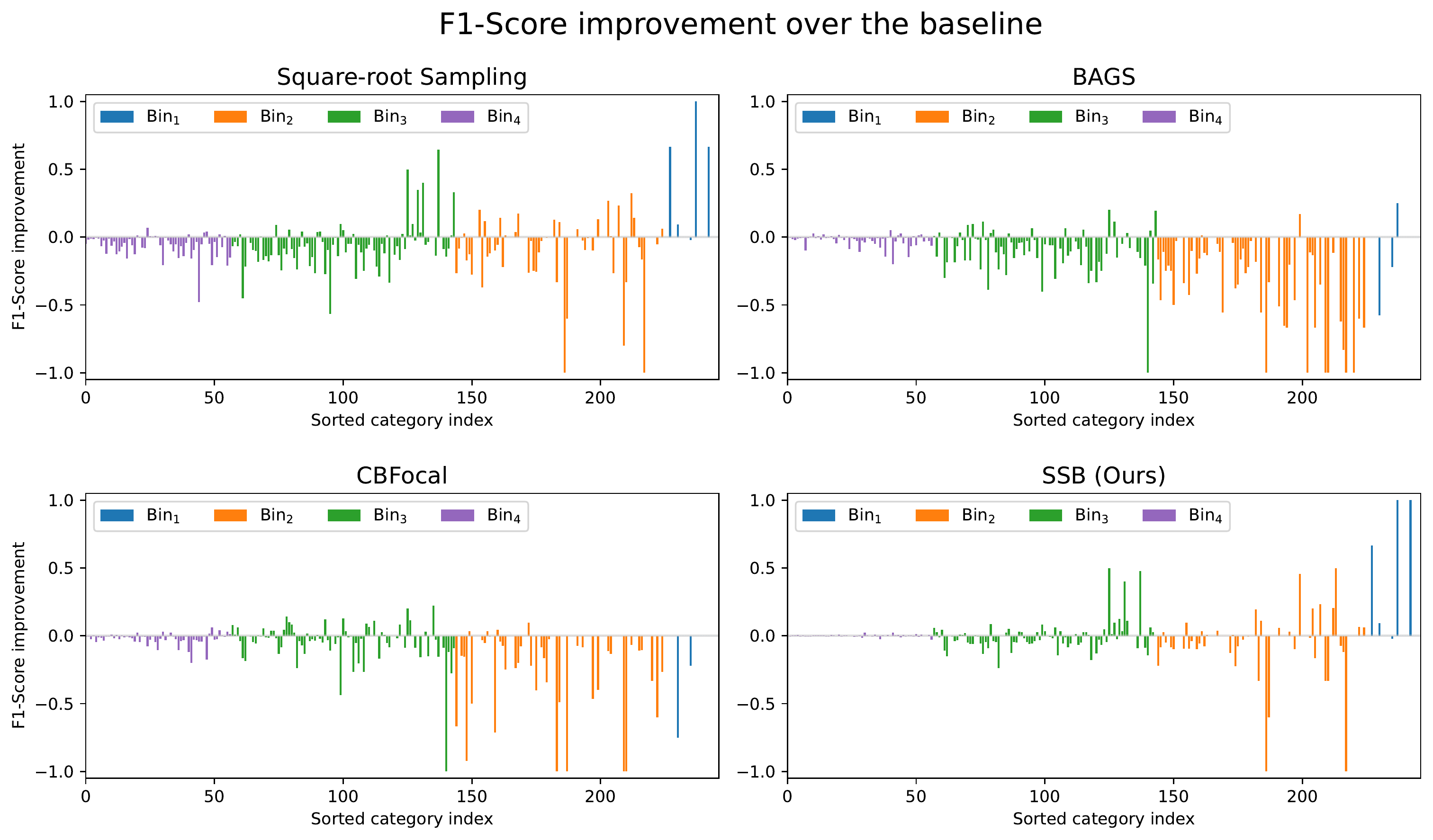}
\end{center}
\vspace{-16pt}
    \caption{Swin-S}
    \label{fig:wcs_fullimage_f1_swin}
\end{subfigure}

   \caption{Breakdown of F1-score improvement per class in the WCS dataset for the long-tail visual recognition methods over the baseline training procedure. For ResNet-50, our proposed SSB is the most consistent method to increase F1-score thoughout the classes. Similar behavior occurs for MobileNetV3-Large and EfficientNetV2-B2 models (not included in the graph). However, for the Swin-S architecture, all the evaluated methods degraded the F1-score, but SSB is the one that lowers the F1-score the least.}
\label{fig:fullimage_f1_increment}
\end{figure*}

In terms of the macro-averaged F1-score, SSB achieved a score that was higher 
than the other methods for most datasets and architectures, as shown in 
\autoref{fig:fullimage_f1}. For all the datasets, the Swin-S trained with the 
baseline settings attained the best macro F1-score among all architectures, with 
Swin-S trained with SSB having the second-best result. 
\autoref{fig:wcs_fullimage_f1_resnet} shows the increment/decrement of the 
F1-score class-wise over the baseline for ResNet-50 in the WCS dataset after 
applying each long-tail method. SSB is the method with the most consistent 
increment in terms of F1-score across classes for ResNet-50. 
\autoref{fig:wcs_fullimage_f1_swin} shows that the decrease in F1-score for the 
Swin-S is generalized over all classes, with SSB being the method that lowers 
the F1-score the least.

\subsection{Comparison between methods when using cropped images}

In this experiment, an assessment was made to see whether using an object 
detector before classifying an animal can improve the recognition score for the 
tail classes. The same settings from the previous experiment were used, except 
for using MegaDetectorV4 to generate cropped regions around animals as inputs, 
though reverting back to the whole image when the confidence score was less than 
0.6. For the pre-cropping experiments, the Snapshot Serengeti dataset was not 
used due to the time required to generate bounding boxes with MegaDetectorV4 and 
run the experiments with the large quantity of images available.

\begin{figure*}[h!]
\centering
 \begin{center}
\includegraphics[width=\textwidth]{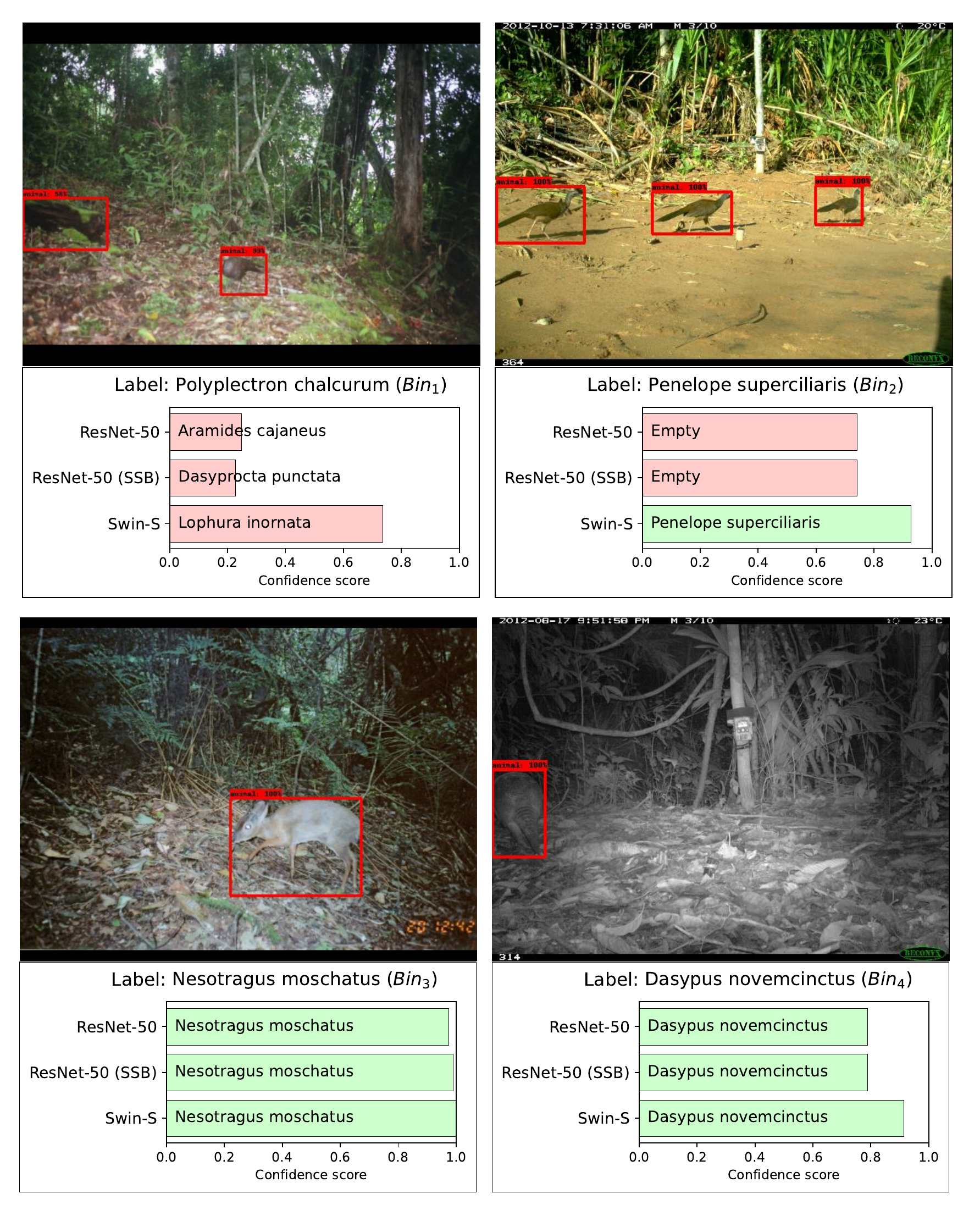}
\end{center}
   \caption{Sample classification results for the baseline ResNet-50, ResNet-50 with the proposed SSB, and Swin-S. One picture was randomly picked from each bin of categories from the WCS dataset. Bounding boxes were generated using MegaDetectorV4.}
\label{fig:image_preds}
\end{figure*}

As shown in \autoref{table:bbox_results}, there was a general accuracy and macro 
F1-score increase when the models were trained using cropped images. However, 
these results follow the same pattern observed when using the entire image as 
the input. Specifically, square-root sampling attained the highest improvements 
for the tail classes at the cost of reducing performance for the head classes. 
Our proposed SSB was able to improve the tail classes’ accuracy with minimal 
accuracy loss for the head classes. In addition, it attained the best macro 
F1-score in most of the cases. Lastly, \autoref{fig:image_preds} shows some 
sample predictions for pictures randomly picked from the WCS dataset with 
bounding boxes generated by MegaDetectorV4.

\begin{landscape}

\begin{table*}[h!]
\footnotesize
\caption{Comparison between SSB accuracy with different methods for long-tail classification of different models and datasets when using the entire image as the input. Bold and underlined values, respectively, indicate the best and second-best result for each model and dataset combination. $Acc_i$ is the top-1 accuracy for the $Bin_i$. As some category bins do not occur in some of the datasets, in this case, the corresponding column was removed.}
\label{table:fullimage_results}
\resizebox{1.6\textwidth}{!}{
\begin{tabular}{ll|lllll|llll|lll|llll}
\hline
\multicolumn{1}{l}{}   & \multicolumn{1}{l|}{} &
\multicolumn{5}{c|}{\textbf{WCS}}

 & \multicolumn{4}{c|}{\textbf{Snapshot Serengeti}}

  &
\multicolumn{3}{c|}{\textbf{Caltech CT}}

  & \multicolumn{4}{c}{\textbf{Wellington CT}}

                                                               \\
\multicolumn{1}{l}{\textbf{Model}} & \multicolumn{1}{l|}{\textbf{Training}} & 
\multicolumn{1}{c}{\textbf{$Acc_1$}} & \multicolumn{1}{c}{\textbf{$Acc_2$}} & 
\multicolumn{1}{c}{\textbf{$Acc_3$}} & \multicolumn{1}{c}{\textbf{$Acc_4$}} & 
\multicolumn{1}{c|}{\textbf{$Acc_{all}$}} &
\multicolumn{1}{c}{\textbf{$Acc_2$}} & \multicolumn{1}{c}{\textbf{$Acc_3$}} & 
\multicolumn{1}{c}{\textbf{$Acc_4$}} &
\multicolumn{1}{c|}{\textbf{$Acc_{all}$}}&
\multicolumn{1}{c}{\textbf{$Acc_2$}} & \multicolumn{1}{c}{\textbf{$Acc_4$}} & 
\multicolumn{1}{c|}{\textbf{$Acc_{all}$}} &
\multicolumn{1}{c}{\textbf{$Acc_2$}} & \multicolumn{1}{c}{\textbf{$Acc_3$}} & 
\multicolumn{1}{c}{\textbf{$Acc_4$}} & \multicolumn{1}{c}{\textbf{$Acc_{all}$}}
\\ \hline
\multirow{5}{*}{\textbf{ResNet-50}} & \textbf{Baseline} & 0.00 & 13.05 & 32.06 &
\textbf{85.02} & {\ul 81.53} & 2.90 & 44.82 & {\ul 92.96} & {\ul 92.79} & 0.00 &
44.95 & 42.19 & 0.0 & 4.58 & \textbf{75.66} & 72.23 \\
 & \textbf{Sqrt-Samp} & \textbf{16.25} & \textbf{28.94} & \textbf{48.98} & 75.38
& 73.47 & {\ul 6.52} & \textbf{51.58} & 87.29 & 87.16 & 0.00 & 39.59 & 37.16 &
0.0 & {\ul 10.65} & 70.91 & 68.01 \\
 & \textbf{CBFocal} & 5.00 & 14.42 & 40.61 & 83.41 & 80.43 & 0.72 & 38.66 &
92.88 & 92.69 & 0.00 & 44.47 & 41.74 & 0.0 & 4.84 & 75.12 & 71.73 \\
 & \textbf{BAGS} & \textbf{16.25} & 14.19 & 34.84 & 84.43 & 81.14 &
\textbf{9.42} & 41.27 & \textbf{93.12} & \textbf{92.94} & \textbf{1.56} &
\textbf{48.04} & \textbf{45.19} & 0.0 & \textbf{25.92} & 75.00 & \textbf{72.64}
\\
 & \textbf{SSB (Ours)} & {\ul 15.00} & {\ul 21.79} & {\ul 40.97} & {\ul 84.85} &
\textbf{81.91} & 5.07 & {\ul 48.16} & 92.91 & 92.75 & {\ul 0.03} & {\ul 46.09} &
{\ul 43.26} & 0.0 & 5.51 & {\ul 75.62} & {\ul 72.24} \\ \hline
\multirow{5}{*}{\textbf{MBNetV3}} & \textbf{Baseline} & 0.00 & 11.92 & 36.52 &
\textbf{84.24} & {\ul 80.98} & 2.17 & 29.78 & \textbf{92.73} & {\ul 92.52} &
0.00 & \textbf{46.14} & \textbf{43.31} & 0.0 & 5.70 & \textbf{75.65} & {\ul
72.28} \\
 & \textbf{Sqrt-Samp} & {\ul 7.50} & \textbf{28.26} & \textbf{46.85} & 76.71 &
74.61 & {\ul 4.35} & \textbf{43.95} & 87.62 & 87.47 & \textbf{0.16} & 37.76 &
35.45 & 0.0 & {\ul 21.68} & 70.45 & 68.10 \\
 & \textbf{CBFocal} & 0.00 & 9.19 & 36.98 & 79.41 & 76.42 & 0.72 & 29.78 & 92.44
& 92.23 & 0.00 & 41.27 & 38.74 & 0.0 & 4.58 & 74.94 & 71.55 \\
 & \textbf{BAGS} & \textbf{10.00} & 10.44 & 28.68 & {\ul 84.09} & 80.48 &
\textbf{5.07} & 28.78 & 92.19 & 91.97 & 0.00 & 43.96 & 41.26 & 0.0 &
\textbf{23.41} & 74.02 & 71.58 \\
 & \textbf{SSB (Ours)} & 1.25 & {\ul 20.32} & {\ul 40.43} & 83.99 &
\textbf{81.04} &
3.62 & {\ul 36.97} & {\ul 92.72} & \textbf{92.53} & {\ul 0.11} & {\ul 45.93} &
{\ul 43.12} & 0.0 & 13.54 & {\ul 75.34} & \textbf{72.36} \\ \hline
\multirow{5}{*}{\textbf{EffV2-B2}} & \textbf{Baseline} & 0.00 & 10.22 & 32.31 &
86.35 & 82.75 & 2.90 & 53.71 & 93.82 & 93.68 & {\ul 0.08} & 56.36 & 52.91 & 0.0
& 5.70 & {\ul 77.11} & 73.67 \\
 & \textbf{Sqrt-Samp} & \textbf{10.00} & \textbf{22.36} & \textbf{50.34} & 78.24
& 76.12 & {\ul 6.52} & \textbf{59.25} & 90.46 & 90.35 & 0.00 & 51.15 & 48.01 &
0.0 & {\ul 8.06} & 74.14 & 70.95 \\
 & \textbf{CBFocal} & 0.00 & 4.20 & 35.55 & \textbf{87.20} & \textbf{83.62} &
0.00 & 46.51 & {\ul 94.51} & {\ul 94.34} & 0.00 & 57.36 & 53.83 & 0.0 & 4.09 &
\textbf{77.31} & {\ul 73.78} \\
 & \textbf{BAGS} & 1.25 & 7.15 & 32.99 & {\ul 87.01} & {\ul 83.37} &
\textbf{7.25} & 51.45 & \textbf{94.77} & \textbf{94.62} & \textbf{0.26} & {\ul
58.88} & {\ul 55.28} & 0.0 & \textbf{30.05} & 76.00 & \textbf{73.79} \\
 & \textbf{SSB (Ours)} & {\ul 7.50} & {\ul 18.39} & {\ul 42.39} & 86.18 & 83.17
& 5.80
& {\ul 55.96} & 93.81 & 93.67 & {\ul 0.08} & \textbf{59.82} & \textbf{56.15} &
0.0 & 6.75 & 77.01 & 73.62 \\ \hline
\multirow{5}{*}{\textbf{Swin-S}} & \textbf{Baseline} & {\ul 8.75} & 37.12 &
62.99 & \textbf{90.86} & \textbf{88.76} & 10.87 & 60.42 & \textbf{95.09} &
\textbf{94.97} & \textbf{4.17} & {\ul 68.94} & {\ul 64.96} & 0.0 & 7.61 &
\textbf{78.94} & {\ul 75.51} \\
 & \textbf{Sqrt-Samp} & \textbf{16.25} & \textbf{44.84} & \textbf{67.22} & 83.49
& 82.13 & 13.04 & \textbf{64.85} & 89.08 & 88.99 & 1.08 & 64.95 & 61.02 & 0.0 &
{\ul 13.09} & 76.54 & 73.48 \\
 & \textbf{CBFocal} & 0.00 & 25.09 & 56.57 & 90.32 & 87.78 & 4.35 & 53.71 &
94.97 & 94.82 & {\ul 3.78} & \textbf{69.61} & \textbf{65.57} & 0.0 & 5.44 &
78.49 & 74.98 \\
 & \textbf{BAGS} & 6.25 & 21.68 & 52.33 & 90.08 & 87.32 & \textbf{15.22} & 52.32
& 94.81 & 94.67 & 3.67 & 66.01 & 62.18 & 0.0 & \textbf{19.65} & 78.62 &
\textbf{75.78} \\
 & \textbf{SSB (Ours)} & \textbf{16.25} & {\ul 40.18} & {\ul 64.91} & {\ul
90.65} &
{\ul 88.69} & {\ul 11.59} & {\ul 61.94} & {\ul 95.05} & {\ul 94.93} & 1.08 &
67.57 & 63.49 & 0.0 & 7.16 & {\ul 78.86} & 75.41
\\\hline
\end{tabular}}
\end{table*}

\end{landscape}

\begin{landscape}

\begin{table*}
\caption{Results attained by different methods and models on different datasets when using the bounding box with the highest confidence as input. Bounding boxes were generated by MegaDetectorV4. Bold and underlined values, respectively, indicate the best and second-best result for each model and dataset combination.}
\label{table:bbox_results}
\setlength\tabcolsep{2pt}
\resizebox{1.6\textwidth}{!}{
\begin{tabular}{ll|llllll|llll|lllll}
\hline
\multicolumn{1}{c}{\textbf{}} & \multicolumn{1}{c|}{\textbf{}} &
\multicolumn{6}{c|}{\textbf{WCS}} & \multicolumn{4}{c|}{\textbf{Caltech CT}} &
\multicolumn{5}{c}{\textbf{Wellington CT}} \\
\multicolumn{1}{c}{\textbf{Model}} & \multicolumn{1}{c|}{\textbf{Training}} &
\multicolumn{1}{c}{\textbf{$Acc_1$}} & \multicolumn{1}{c}{\textbf{$Acc_2$}} &
\multicolumn{1}{c}{\textbf{$Acc_3$}} & \multicolumn{1}{c}{\textbf{$Acc_4$}} &
\multicolumn{1}{c}{\textbf{$Acc_all$}} & \multicolumn{1}{c|}{\textbf{$F_1^m$}} &
\multicolumn{1}{c}{\textbf{$Acc_2$}} & \multicolumn{1}{c}{\textbf{$Acc_4$}} &
\multicolumn{1}{c}{\textbf{$Acc_all$}} & \multicolumn{1}{c|}{\textbf{$F_1^m$}} &
\multicolumn{1}{c}{\textbf{$Acc_2$}} & \multicolumn{1}{c}{\textbf{$Acc_3$}} &
\multicolumn{1}{c}{\textbf{$Acc_4$}} & \multicolumn{1}{c}{\textbf{$Acc_all$}} &
\multicolumn{1}{c}{\textbf{$F_1^m$}} \\ \hline
\multirow{5}{*}{\textbf{ResNet-50}} & \textbf{Baseline} & 0.00 & 21.68 & 52.94 &
90.61 & 87.85 & 0.429 & 0.00 & 60.59 & 56.87 & 0.423 & 0.0 & 8.18 & {\ul 78.52}
& 75.14 & 0.408 \\
 & \textbf{Sqrt-Samp} & {\ul 20.00} & \textbf{43.47} & \textbf{67.19} & 85.79 &
84.28 & 0.473 & \textbf{0.24} & 62.62 & 58.79 & 0.448 & 0.0 & {\ul 22.21} &
75.32 & 72.76 & \textbf{0.431} \\
 & \textbf{CBFocal} & 3.75 & 26.33 & 61.67 & {\ul 90.98} & \textbf{88.65} &
0.468 & 0.00 & 64.32 & 60.37 & 0.458 & 0.0 & 10.77 & 78.47 & 75.21 & {\ul 0.420}
\\
 & \textbf{BAGS} & \textbf{26.25} & 23.16 & 53.33 & \textbf{91.00} & {\ul 88.28}
& 0.414 & {\ul 0.13} & \textbf{67.01} & \textbf{62.91} & {\ul 0.464} & 0.0 &
\textbf{36.98} & 77.92 & \textbf{75.95} & 0.409 \\
 & \textbf{SSB (Ours)} & 17.50 & {\ul 35.41} & {\ul 61.95} & 90.39 & 88.25 &
\textbf{0.501} & 0.11 & {\ul 65.79} & {\ul 61.76} & \textbf{0.467} & 0.0 & 12.23
& \textbf{78.73} & {\ul 75.53} & 0.416 \\ \hline
\multirow{5}{*}{\textbf{MBNetV3}} & \textbf{Baseline} & 0.00 & 26.11 & 55.93 &
\textbf{91.34} & {\ul 88.72} & 0.434 & 0.00 & 64.80 & 60.82 & {\ul 0.471} & 0.0
& 11.89 & {\ul 80.02} & 76.74 & 0.434 \\
 & \textbf{Sqrt-Samp} & \textbf{26.25} & \textbf{42.22} & \textbf{64.41} & 87.30
& 85.57 & {\ul 0.458} & {\ul 0.29} & 57.41 & 53.91 & 0.436 & 0.0 & {\ul 23.89} &
75.67 & 73.17 & \textbf{0.441} \\
 & \textbf{CBFocal} & 1.25 & 19.64 & 56.86 & 91.04 & 88.39 & 0.418 & 0.13 &
61.29 & 57.53 & 0.468 & 0.0 & 11.44 & 79.30 & 76.03 & 0.422 \\
 & \textbf{BAGS} & 13.75 & 22.47 & 45.07 & 90.13 & 87.07 & 0.373 & \textbf{0.42}
& {\ul 65.10} & {\ul 61.12} & 0.461 & 0.0 & \textbf{38.60} & 79.31 &
\textbf{77.34} & 0.403 \\
 & \textbf{SSB (Ours)} & {\ul 21.25} & {\ul 35.19} & {\ul 60.35} & {\ul 91.13} &
\textbf{88.88} & \textbf{0.498} & 0.11 & \textbf{66.59} & \textbf{62.50} &
\textbf{0.489} & 0.0 & 15.98 & \textbf{80.30} & {\ul 77.20} & {\ul 0.438} \\
\hline
\multirow{5}{*}{\textbf{EffV2-B2}} & \textbf{Baseline} & 0.00 & 19.64 & 48.13 &
90.46 & 87.46 & 0.383 & {\ul 0.11} & 61.34 & 57.58 & 0.424 & 0.0 & 8.40 & {\ul
79.61} & 76.18 & 0.427 \\
 & \textbf{Sqrt-Samp} & \textbf{17.50} & \textbf{35.19} & \textbf{65.09} & 88.37
& 86.49 & {\ul 0.469} & 0.08 & 60.33 & 56.63 & 0.437 & 0.0 & {\ul 23.33} & 76.86
& 74.28 & \textbf{0.452} \\
 & \textbf{CBFocal} & 0.00 & 10.33 & 53.58 & \textbf{91.34} & \textbf{88.40} &
0.392 & 0.00 & 64.72 & 60.74 & {\ul 0.468} & 0.0 & 12.53 & 79.39 & 76.17 & 0.430
\\
 & \textbf{BAGS} & 0.00 & 13.17 & 51.87 & {\ul 91.24} & {\ul 88.26} & 0.384 &
\textbf{0.61} & \textbf{68.03} & \textbf{63.89} & \textbf{0.478} & 0.0 &
\textbf{44.22} & 79.15 & \textbf{77.46} & 0.429 \\
 & \textbf{SSB (Ours)} & {\ul 16.25} & {\ul 30.31} & {\ul 60.28} & 90.38 & 88.10
& \textbf{0.474} & 0.08 & {\ul 67.10} & {\ul 62.98} & 0.465 & 0.0 & 12.27 &
\textbf{79.80} & {\ul 76.55} & {\ul 0.439} \\ \hline
\multirow{5}{*}{\textbf{Swin-S}} & \textbf{Baseline} & 16.25 & 53.35 & 78.20 &
{\ul 93.63} & \textbf{92.28} & \textbf{0.637} & {\ul 10.32} & {\ul 70.49} & {\ul
66.80} & \textbf{0.597} & 0.0 & 11.25 & {\ul 80.14} & {\ul 76.82} & 0.503 \\
 & \textbf{Sqrt-Samp} & \textbf{25.00} & \textbf{62.09} & \textbf{79.37} & 90.18
& 89.22 & 0.586 & 2.90 & 67.84 & 63.85 & 0.524 & 0.0 & \textbf{19.05} & 78.42 &
75.56 & {\ul 0.504} \\
 & \textbf{CBFocal} & 8.75 & 39.95 & 75.24 & \textbf{93.66} & {\ul 91.97} & {\ul
0.599} & \textbf{11.40} & 69.41 & 65.84 & {\ul 0.596} & 0.0 & 9.41 & 79.25 &
75.88 & 0.476 \\
 & \textbf{BAGS} & 8.75 & 33.83 & 69.93 & 93.52 & 91.52 & 0.511 & 8.29 &
\textbf{73.86} & \textbf{69.84} & 0.560 & 0.0 & {\ul 18.64} & 79.36 & 76.44 &
0.447 \\
 & \textbf{SSB (Ours)} & {\ul 23.75} & {\ul 58.34} & {\ul 78.55} & 93.53 &
\textbf{92.28} & \textbf{0.637} & 3.27 & 69.52 & 65.45 & 0.576 & 0.0 & 12.34 &
\textbf{80.75} & \textbf{77.46} & \textbf{0.514} \\ \hline
\end{tabular}}
\end{table*}

\end{landscape}

\subsection{Comparing SSB to one-stage training
strategies}

In this experiment, the performance of single-stage training methods was 
evaluated by applying re-weighting (CBFocal) and re-sampling (square-root 
sampling). For both methods, cropping was also used before classification, with 
bounding boxes being generated by MegaDetectorV4. 
\autoref{table:onestage_results} shows the results and compares them to the 
baseline training procedure and to our proposed two-stage SSB. The results show 
that square-root sampling presents similar behavior as observed in the two-stage 
training, i.e., improving the performance of the tail classes, but reducing the 
performance of the head classes. On the other hand, CBFocal obtained a great 
improvement over its results in two-stage training, and achieved the best 
overall accuracy in most cases. However, this improvement occurred mostly in the 
head classes. One-stage training methods also fail to improve the classification 
performance of the tail classes from the Caltech and Wellington datasets. 
Considering the objective of improving accuracy of the tail classes, our 
proposed SSB still represents the best trade-off between improving accuracy of 
the tail classes and reducing performance in the head classes.

\begin{table*}[htp]
\footnotesize
\caption{Comparing the results of one-stage training using square-root re-sampling and CBFocal loss with the baseline training procedure and our proposed SSB. Bold and underlined values, respectively, indicate the best and second-best result for each model and dataset combination.}
\label{table:onestage_results}
\setlength\tabcolsep{2pt}
\resizebox{\textwidth}{!}{
\begin{tabular}{ll|llllll|llll|lllll}
\hline
\multicolumn{1}{c}{\textbf{}} & \multicolumn{1}{c|}{\textbf{}} &
\multicolumn{6}{c|}{\textbf{WCS}} & \multicolumn{4}{c|}{\textbf{Caltech CT}} &
\multicolumn{5}{c}{\textbf{Wellington CT}} \\
\multicolumn{1}{c}{\textbf{Model}} & \multicolumn{1}{c|}{\textbf{Training}} &
\multicolumn{1}{c}{\textbf{$Acc_1$}} & \multicolumn{1}{c}{\textbf{$Acc_2$}} &
\multicolumn{1}{c}{\textbf{$Acc_3$}} & \multicolumn{1}{c}{\textbf{$Acc_4$}} &
\multicolumn{1}{c}{\textbf{$Acc_all$}} & \multicolumn{1}{c|}{\textbf{$F_1^m$}} &
\multicolumn{1}{c}{\textbf{$Acc_2$}} & \multicolumn{1}{c}{\textbf{$Acc_4$}} &
\multicolumn{1}{c}{\textbf{$Acc_all$}} & \multicolumn{1}{c|}{\textbf{$F_1^m$}} &
\multicolumn{1}{c}{\textbf{$Acc_2$}} & \multicolumn{1}{c}{\textbf{$Acc_3$}} &
\multicolumn{1}{c}{\textbf{$Acc_4$}} & \multicolumn{1}{c}{\textbf{$Acc_all$}} &
\multicolumn{1}{c}{\textbf{$F_1^m$}} \\ \hline
\multirow{4}{*}{\textbf{ResNet-50}} & \textbf{Baseline} & 0.00 & 21.68 & 52.94 &
{\ul 90.61} & 87.85 & 0.429 & 0.00 & 60.59 & 56.87 & 0.423 & 0.0 & 8.18 & 78.52
& 75.14 & 0.408 \\
 & \textbf{Sqrt-Samp} & \textbf{20.00} & \textbf{45.86} & \textbf{71.61} & 84.07
& 82.90 & {\ul 0.530} & \textbf{0.90} & 64.87 & 60.94 & {\ul 0.529} & 0.0 & {\ul
13.58} & 77.29 & 74.22 & {\ul 0.458} \\
 & \textbf{CBFocal} & 15.00 & {\ul 35.98} & {\ul 67.58} & \textbf{92.13} &
\textbf{90.15} & \textbf{0.538} & 0.05 & \textbf{71.67} & \textbf{67.27} &
\textbf{0.541} & 0.0 & \textbf{15.00} & \textbf{79.07} & \textbf{75.99} &
\textbf{0.460} \\
 & \textbf{SSB (Ours)} & {\ul 17.50} & 35.41 & 61.95 & 90.39 & {\ul 88.25} &
0.501 & {\ul 0.11} & {\ul 65.79} & {\ul 61.76} & 0.467 & 0.0 & 12.23 & {\ul
78.73} & {\ul 75.53} & 0.416 \\ \hline
\multirow{4}{*}{\textbf{MBNetV3}} & \textbf{Baseline} & 0.00 & 26.11 & 55.93 &
{\ul 91.34} & 88.72 & 0.434 & 0.00 & {\ul 64.80} & {\ul 60.82} & 0.471 & 0.0 &
11.89 & 80.02 & 76.74 & 0.434 \\
 & \textbf{Sqrt-Samp} & \textbf{22.50} & \textbf{48.13} & \textbf{71.75} & 84.85
& 83.67 & \textbf{0.504} & {\ul \textbf{1.08}} & 58.20 & 54.69 & {\ul 0.487} &
0.0 & 14.55 & 76.35 & 73.37 & \textbf{0.459} \\
 & \textbf{CBFocal} & 3.75 & 24.86 & {\ul 61.99} & \textbf{91.69} &
\textbf{89.31} & 0.463 & 0.08 & 64.23 & 60.29 & 0.485 & 0.0 & {\ul 15.64} & {\ul
80.11} & {\ul 77.00} & {\ul 0.452} \\
 & \textbf{SSB (Ours)} & {\ul 21.25} & {\ul 35.19} & 60.35 & 91.13 & {\ul 88.88}
& {\ul 0.498} & {\ul 0.11} & \textbf{66.59} & \textbf{62.50} & \textbf{0.489} &
0.0 & \textbf{15.98} & \textbf{80.30} & \textbf{77.20} & 0.438 \\ \hline
\multirow{4}{*}{\textbf{EffV2-B2}} & \textbf{Baseline} & 0.00 & 19.64 & 48.13 &
{\ul 90.46} & 87.46 & 0.383 & {\ul 0.11} & 61.34 & 57.58 & 0.424 & 0.0 & 8.40 &
79.61 & 76.18 & 0.427 \\
 & \textbf{Sqrt-Samp} & \textbf{26.25} & \textbf{56.64} & \textbf{78.09} & 85.66
& 84.84 & \textbf{0.565} & \textbf{2.61} & 61.37 & 57.76 & {\ul 0.500} & 0.0 &
{\ul 16.54} & 77.07 & 74.15 & {\ul 0.460} \\
 & \textbf{CBFocal} & 0.00 & 13.85 & 60.24 & \textbf{92.37} & \textbf{89.72} &
0.410 & 0.00 & \textbf{71.39} & \textbf{67.00} & \textbf{0.529} & 0.0 &
\textbf{18.60} & \textbf{80.75} & \textbf{77.76} & \textbf{0.471} \\
 & \textbf{SSB (Ours)} & {\ul 16.25} & {\ul 30.31} & {\ul 60.28} & 90.38 & {\ul
88.10} & {\ul 0.474} & 0.08 & {\ul 67.10} & {\ul 62.98} & 0.465 & 0.0 & 12.27 &
{\ul 79.80} & {\ul 76.55} & 0.439 \\ \hline
\multirow{4}{*}{\textbf{Swin-S}} & \textbf{Baseline} & 16.25 & 53.35 & 78.20 &
{\ul 93.63} & {\ul 92.28} & {\ul 0.637} & \textbf{10.32} & {\ul 70.49} & {\ul
66.80} & {\ul 0.597} & 0.0 & 11.25 & 80.14 & 76.82 & 0.503 \\
 & \textbf{Sqrt-Samp} & \textbf{23.75} & 51.87 & \textbf{78.62} & 92.03 & 90.78
& 0.634 & 2.16 & 67.71 & 63.68 & 0.574 & 0.0 & 11.29 & 79.36 & 76.08 & 0.479 \\
 & \textbf{CBFocal} & {\ul 21.25} & {\ul 54.03} & 78.41 & \textbf{93.97} &
\textbf{92.62} & \textbf{0.642} & {\ul 6.76} & \textbf{71.62} & \textbf{67.63} &
\textbf{0.606} & 0.0 & \textbf{13.43} & {\ul 80.56} & {\ul 77.33} &
\textbf{0.524} \\
 & \textbf{SSB (Ours)} & \textbf{23.75} & \textbf{58.34} & {\ul 78.55} & 93.53 &
{\ul 92.28} & {\ul 0.637} & 3.27 & 69.52 & 65.45 & 0.576 & 0.0 & {\ul 12.34} &
\textbf{80.75} & \textbf{77.46} & {\ul 0.514} \\ \hline
\end{tabular}}
\end{table*}

\section{Discussion}

The long-tail distribution is a characteristic of real-world tasks, such as 
visual recognition of species in camera-trap images, in which some classes 
concentrate the majority of available instances and a large number of classes 
have just a small quantity of samples. However, as indicated by 
\citet{schneider2020three}, computer vision models generally often perform 
poorly on recognizing species in camera-trap images with fewer training 
instances (tail classes). In this work, a simple and effective framework called 
Square-root Sampling Branch (SSB) is proposed, which can improve the performance 
in tail classes with a minimal percentage of accuracy loss in the head classes. 
The proposed framework was systematically evaluated against state-of-the-art 
methods for handling the long-tail distribution issue in four different 
camera-trap datasets and four families of computer vision models. Our 
experiments show that our proposed framework presents competitive results with 
the methods in the literature, achieving the best trade-off between increasing 
the performance of the tail classes and reducing the head classes’ accuracy in 
most cases.

Comparing our improved baseline training procedure with the literature, 
ResNet-50 achieved an overall accuracy of 92.79\% for the Snapshot Serengeti 
dataset using the full image, while the same architecture achieved an accuracy 
of 93.6\% in the work conducted by \citet{norouzzadeh2018automatically} and 
approximately 85\% in that of \citet{villa2017towards}. However, while these 
works use burst or random split, in our methodology we split the training/test 
partitions based on locations, which tends to achieve a significantly lower 
accuracy, as demonstrated for other datasets by \citet{beery2018recognition}, 
\citet{schneider2020three}, and \citet{tabak2019machine}, in which the 
performance of models on untrained locations drops from 15\% to 25\% in accuracy 
in terms of absolute values. Furthermore, the categories and test sets used are 
different among the works, thus making the results not directly comparable. The 
same occurs when comparing our results for the Caltech CT dataset with the 
results from \citet{beery2018recognition}. The lack of well-defined benchmarks 
for visual recognition of animal species in camera-trap images makes it 
difficult to compare the results of different methods across works. Although 
most of the applications have their specificities and requirements regarding the 
data preparation or even the target dataset, it would be important to apply them 
using a standardized benchmark as well. However, designing such a benchmark
dataset is challenging due to the wide variety of factors that need to be
considered, including taxonomic coverage, biogeography, habitat and
climate conditions, as well as the tasks to be included, such as species
classification, long-tail visual recognition, animal detection, counting, and
behavior identification. Therefore, it is recommended that the research
community work collaboratively towards developing standardized benchmarks for
automatic information extraction from camera-trap images, which would facilitate
comparisons between different methods and ultimately lead to advancements in
this field.

In general, the method that most improved the performance for the tail classes 
over the baseline was the square-root sampling. However, this method strongly 
degrades the head classes’ accuracy. This behavior may be due to the high 
imbalance of the datasets, leading the square-root sampling to cause a high 
undersampling of the majority classes. Our proposed SSB can improve the accuracy 
for the tail classes more than CBFocal and BAGS for most of the cases, but, 
unlike square-root sampling, the degradation caused to the head classes is 
minimal, representing the best trade-off among all the evaluated methods. SSB 
also represents the best performance in terms of macro F1-score for most of the 
cases when compared with the two-stage methods. On the other hand, our 
experiments using one training stage show that CBFocal achieved the best overall 
accuracy for most of the cases, primarily because of its superior performance in
the head classes. This CBFocal performance improvement may indicate that
pretraining using softmax cross-entropy does not generate a good representation
for fine-tuning using Focal loss, or that the learning of representation and
classification is not well decoupled when using this loss function. However, our
experiments were not designed to assess the performance based on representation
(feature extractor) training strategies. Furthermore, \citet{Kang2020Decoupling}
does not evaluate the performance of re-weighting methods when proposing the
two-stage training approach. Although this approach was not explored in our
experiments, a potential alternative could be to combine the outcomes of
square-root resampling and CBFocal, which respectively performed better on the
tail and head classes. However, this approach would require running inference
using two models, effectively doubling the computational cost. In contrast, our
proposed SSB only needs to be run once and is still competitive.

Looking into the model’s performance, Swin Transformer obtained the best results 
considering all the datasets, which was expected since this model has more 
parameters than the other selected models. It is worth noting that the Swin-S 
that was trained using the baseline settings attained an excellent performance 
even for the tail classes, reaching the best mean F1-score in most of the 
datasets. This superior performance of vision transformers in long-tail 
classification corroborates with what has been stated in the literature, as in 
the classification of fungi \citep{picek2022danish}, for instance.

Regarding the question of whether using cropping before classification can 
improve model performance, when comparing the results of the first and the 
second series of experiments, it is possible to note a significant increase in 
the overall accuracy when applying cropping. This improvement is more relevant 
for the WCS and Caltech datasets, in which, for some cases, there was an 
improvement of 10\% and 15\%, respectively. This performance improvement is 
consistent with the results in the literature, for example, in 
\citep{beery2018recognition}, there was a 20\% increase in the accuracy when 
cropping was used before classification. The use of an object detector as a 
preprocessing step before species identification is also part of the methodology 
of \citet{schneider2018deep} and \citet{norouzzadeh2021deep}; however, its 
impact on the performance is not evaluated. Looking at the binned accuracy 
results, a performance increase is observed for all bins of the classes, except 
for the tail classes ($Bin_2$) for the Caltech CT and Wellington CT datasets, in 
which all models and methods fail to learn them. The lowest performance 
increases due to the use of cropping before classification were observed for the 
Swin-S model, with an improvement varying from 1\% to 4\%. It is worth noting 
that the accuracies for Swin-S using the full images are already very high when 
compared to the other models. This can be explained by the vision transformer’s 
attention mechanism that can allow it to focus on the animal even when using the 
full image.

Moreover, our experiments did not explain why the models were unable to learn to
recognize the tail classes in the smaller datasets. The Caltech and Wellington
datasets, with only 19 and 14 classes respectively, had a limited number of tail
classes (4 and 1, respectively). It is possible that these tail classes were
ignored due to the relatively low number of training samples available. To
investigate this further, future studies could use the WCS dataset and
progressively remove classes to evaluate the impact on model performance. This
would also provide insights into the minimum number of training images per
class needed to see an improvement in performance when using the evaluated
methods.

One limitation to our study is that only the WCS dataset has classes belonging 
to all bins of classes in regards to the number of training instances. Another
limitation is that, during our experiments, there was an effort to keep our
settings as similar as possible across the methods and models in order to make a
fair comparison; therefore, our hyperparameters are suboptimal and better tuning
would lead to improvements.

\section{Conclusions}

Although rare or endangered species may be the classes of interest for projects 
using camera traps for monitoring wildlife, they are often neglected when 
developing deep-learning models for automatic species recognition, mostly 
because of the difficulty involved in learning classes with few training 
instances from long-tail datasets. In this work, a simple and effective 
framework called Square-root Sampling Branch (SSB) is proposed, which uses a 
multi-branch approach to improve the performance recognition of tail classes of 
long-tail camera-trap datasets with minimal loss in the performance recognition 
of the head classes. Our proposed method was systematically evaluated against 
square-root sampling, CBFocal loss, and Balanced Group Softmax (BAGS) 
considering four camera-trap datasets: Snapshot Serengeti, WCS, Caltech CT, and 
Wellington CT. Our results showed that our proposed approach achieved the best 
trade-off between improving accuracy for the tail classes and decreasing the 
accuracy of the head classes, and is competitive with the state-of-the-art 
methods. It was also found that Swin-S performed better for most of the datasets 
even without applying any additional method for handling imbalance, and achieved 
an overall accuracy of 88.76\% for the WCS dataset and 94.97\% for the Snapshot 
Serengeti, when considering location-based splitting.

Despite the fact that our proposed framework improves the recognition of tail 
classes, our experiments clearly show that it is necessary to develop more 
sophisticated methods to handle the long-tail distribution in the recognition of 
species in camera-trap images. Specifically for the tail classes ($Bin_1$ and 
$Bin_2$), for which the results were extremely low, and especially for the small 
datasets, one could evaluate few-shot learning methods. Another issue that was 
discussed is the lack of standardized benchmarks for evaluating the performance 
of long-tail visual recognition of animal species when comparing new methods. It 
is hoped that our well-defined methodology for dataset preprocessing and 
training procedure helps establish good baselines for other researchers. In 
future works, other methods can be explored such as progressive-balanced 
sampling, Ally Complementary Experts, and Rebalanced Mixup (Remix). Additional 
training tricks, such as label smoothing and mixup, can also be performed, but 
they must be used with caution because one trick may affect the other.

\section*{Acknowledgements}

We would like to thank Coordenação de Aperfeiçoamento de Pessoal de Nível Superior– CAPES/PROAP. We also thank Fundação de Amparo à Pesquisa do Estado do Amazonas – FAPEAM/POSGRAD 2021. The present work is the result of the research and development (R\&D) project No. 001/2020, between the Federal University of Amazonas and FAEPI, Brazil, which has funding from Samsung, and uses resources from the Informatics Law for the Western Amazon (Federal Law nº 8.387/1991), and its disclosure is in accordance with article 39 of Decree No. 10.521/2020. The funders had no role in study design, data collection and analysis, decision to publish, or preparation of the manuscript.


\bibliography{citations}

\end{document}